\def\year{2022}\relax
\documentclass[letterpaper]{article} % DO NOT CHANGE THIS
\usepackage{aaai22}  % DO NOT CHANGE THIS
\usepackage{times}  % DO NOT CHANGE THIS
\usepackage{helvet}  % DO NOT CHANGE THIS
\usepackage{courier}  % DO NOT CHANGE THIS
\usepackage[hyphens]{url}  % DO NOT CHANGE THIS
\usepackage{graphicx} % DO NOT CHANGE THIS
\usepackage{natbib}  % DO NOT CHANGE THIS AND DO NOT ADD ANY OPTIONS TO IT
\usepackage{caption} % DO NOT CHANGE THIS AND DO NOT ADD ANY OPTIONS TO IT
\DeclareCaptionStyle{ruled}{labelfont=normalfont,labelsep=colon,strut=off} % DO NOT CHANGE THIS
\usepackage{algorithm}
\usepackage{algorithmic}

\usepackage{newfloat}
\usepackage{listings}

\usepackage{amsmath}
\usepackage{xspace}
\usepackage{dsfont}
\usepackage{epsfig}
\usepackage{graphicx}
\usepackage{amssymb}
\usepackage{bbm}
\usepackage{color}
\usepackage{float}
\usepackage{multirow}
\usepackage{makecell}
\usepackage{subfigure}

\newcommand{\T}{T\xspace}
\newcommand{\D}{D\xspace}

\newcommand{\A}{x}
\newcommand{\B}{x^v}
\newcommand{\z}{\delta x}

\usepackage{etoolbox}
\makeatletter
\AfterEndEnvironment{algorithm}{\let\@algcomment\relax}
\AtEndEnvironment{algorithm}{\kern2pt\hrule\relax\vskip3pt\@algcomment}
\let\@algcomment\relax
\newcommand\algcomment[1]{\def\@algcomment{\footnotesize#1}}
\renewcommand\fs@ruled{\def\@fs@cfont{\bfseries}\let\@fs@capt\floatc@ruled
  \def\@fs@pre{\hrule height.8pt depth0pt \kern2pt}%
  \def\@fs@post{}%
  \def\@fs@mid{\kern2pt\hrule\kern2pt}%
  \let\@fs@iftopcapt\iftrue}
\makeatother
\floatstyle{ruled}
\newfloat{listing}{tb}{lst}{}
\floatname{listing}{Listing}
\title{VITA: A Multi-Source Vicinal Transfer Augmentation Method for Out-of-Distribution Generalization}

\author {
    Minghui Chen\textsuperscript{\rm 1},
    Cheng Wen\textsuperscript{\rm 2},
    Feng Zheng\textsuperscript{\rm 1}\footnote{Corresponding Author.},
    Fengxiang He\textsuperscript{\rm 3},
    Ling Shao\textsuperscript{\rm 4}
}
\affiliations {
    \textsuperscript{\rm 1} Department of Computer Science and Engineering, Southern University of Science and Technology \\
    \textsuperscript{\rm 2} The University of Sydney 
    \textsuperscript{\rm 3} JD Explore Academy, JD.com Inc \\
    \textsuperscript{\rm 4} National Center for Artificial Intelligence, Saudi Data and Artificial Intelligence Authority, Riyadh, Saudi Arabia \\    
    ming\_hui.chen@outlook.com, cwen6671@uni.sydney.edu.au, f.zheng@ieee.org, hefengxiang@jd.com, ling.shao@ieee.org
}

\renewcommand{\topfraction}{0.9}
\renewcommand{\dbltopfraction}{0.9}
\renewcommand{\textfraction}{0.08}
\renewcommand{\floatpagefraction}{0.8}
\renewcommand{\dblfloatpagefraction}{0.8}
\selectfont

\begin{document}

\maketitle

\begin{abstract}
    Invariance to diverse types of image corruption, such as noise, blurring, or colour shifts, is essential to establish robust models in computer vision. Data augmentation has been the major approach in improving the robustness against common corruptions. However, the samples produced by popular augmentation strategies deviate significantly from the underlying data manifold. As a result, performance is skewed toward certain types of corruption. To address this issue, we propose a multi-source vicinal transfer augmentation (VITA) method for generating diverse on-manifold samples. The proposed VITA consists of two complementary parts: tangent transfer and integration of multi-source vicinal samples. The tangent transfer creates initial augmented samples for improving corruption robustness. The integration employs a generative model to characterize the underlying manifold built by vicinal samples, facilitating the generation of on-manifold samples. Our proposed VITA significantly outperforms the current state-of-the-art augmentation methods, demonstrated in extensive experiments on corruption benchmarks. Code: \url{https://github.com/MinghuiChen43/VITA}. 
\end{abstract}

\section{Introduction}
\label{sec:Introduction}

\begin{figure*}[!t]
    \centering
    \subfigure[Lack of vicinity sampling]{
        \label{fig:benefit_vita.1}
        \begin{minipage}[b]{0.30\textwidth}
            \includegraphics[width=1\textwidth]{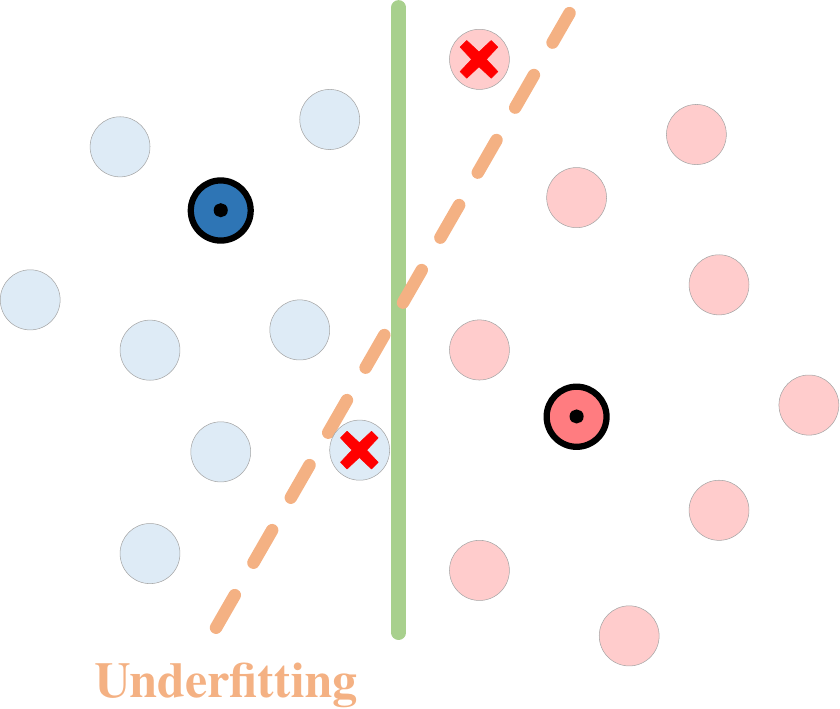}
        \end{minipage}}
    \hspace{8pt}
    \subfigure[Off-manifold samples hurts]{
        \label{fig:benefit_vita.2}
        \begin{minipage}[b]{0.30\textwidth}
            \includegraphics[width=1\textwidth]{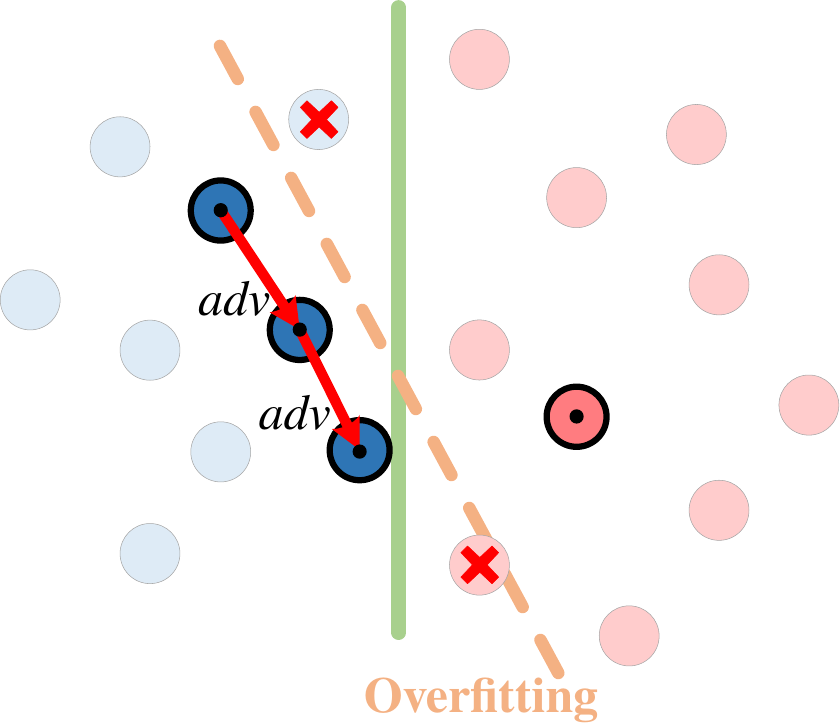}
        \end{minipage}}
    \hspace{8pt}
    \subfigure[Vicinal difference transfer]{
        \label{fig:benefit_vita.3}
        \begin{minipage}[b]{0.30\textwidth}
            \includegraphics[width=1\textwidth]{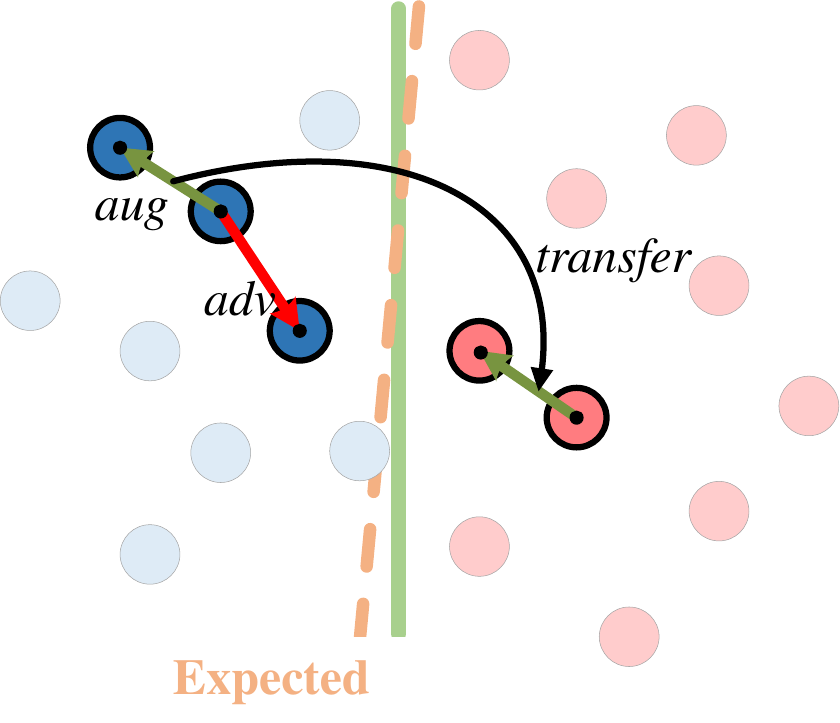}
        \end{minipage}}
    \caption{Diverse on-manifold samples boost performance. \textbf{(a)} shows an underfitting phenomenon caused by the lack of sufficiently diverse vicinal samples. \textbf{(b)} takes a strong adversarial attack as an example to show the performance degradation caused by off-manifold samples. Classifiers tend to overfit these off-manifold samples and fail to construct a proper underlying manifold. \textbf{(c)} demonstrates the benefit of tangent transfer. While maintaining sample variety, it will not generate severe out-of-manifold samples. For a better understanding of samples that depart from the manifold, see our appendix for a three-dimensional schematic diagram.}
    \label{fig:benefit_vita}
\end{figure*}

Existing computer vision systems are not as robust as human vision systems \cite{DBLP:journals/corr/abs-1806-00451, DBLP:conf/iclr/HendrycksD19}.
A human vision system would not confused by a wide range of naturally occurring corruptions, including noise, blurring, and pixelation, as well as some unexpected combinations of them.
However, existing deep models \cite{DBLP:conf/nips/KrizhevskySH12, DBLP:conf/cvpr/XieGDTH17} trained on clean images usually perform substantially worse, confronted with corrupted images \cite{DBLP:conf/nips/GeirhosTRSBW18,DBLP:conf/iclr/HendrycksD19}.
Achieving corruption robustness is a primary objective of a variety of computer vision tasks.
% Achieving {\color{red}this level} of robustness is a primary objective of a variety of computer vision tasks.

The most effective and commonly used method for improving corruption robustness is data augmentation, in which training samples undergo label-preserving transformations \cite{DBLP:conf/icml/DaoGRSSR19}, formalized by the vicinal risk minimization (VRM) principle \cite{DBLP:conf/nips/ChapelleWBV00}.
Employing adversarial examples \cite{DBLP:journals/corr/GoodfellowSS14, DBLP:conf/iclr/MadryMSTV18} or transformation strategies \cite{DBLP:journals/corr/abs-1708-04552, DBLP:conf/iclr/ZhangCDL18, DBLP:conf/iccv/YunHCOYC19} based on human expertise brings limited benefit to corruption robustness.
This is due to the fact that these two methods of augmentation fail to provide sufficiently diverse vicinal samples.
While advanced methods based on generative models \cite{DBLP:conf/iclr/GeirhosRMBWB19, DBLP:conf/eccv/RusakSZB0BB20, DBLP:journals/corr/abs-2006-16241} and combination strategies \cite{DBLP:conf/cvpr/CubukZMVL19, DBLP:conf/iclr/HendrycksMCZGL20} are capable of producing diverse samples, they frequently generate samples that deviate severely from the data manifold.
These samples impair the classifier's ability to estimate the underlying data manifold accurately, resulting in performance degradation and bias towards specific corruptions.

To mitigate this defect in data augmentation, we propose a multi-source vicinal transfer augmentation (VITA) method for generating diverse on-manifold samples.
The proposed VITA is composed of two components: tangent transfer and integration of multi-source vicinal samples.
First, we leverage vicinal differences to approximate the manifold tangents to acquire initial augmented samples.
Subsequent experiments show that these weakly augmented samples effectively improve corruption robustness.
Second, we use a generative model to characterize the underlying data manifold constructed by weakly augmented samples (\textit{e.g.}, samples rotated by 5 degrees) and adversarial examples.
This is to ensure that a diverse set of samples is generated while avoiding significant deviance from the data manifold \cite{DBLP:conf/nips/BengioYAV13}.
Detailed experiments confirm that our VITA can significantly improve corruption robustness and encourage a balanced performance on corrupted datasets.

In summary, our key contributions are as follows:
\begin{itemize}
    \item To address the uneven performance toward various corrupted images, we propose a multi-source vicinal transfer augmentation (VITA) method for generating diverse on-manifold samples.
    \item We introduce tangent transfer that enforces the local invariance of the classifier, which facilitates the discovery of shared structures in the tangent planes.
    \item We design an integration module of multi-source vicinal samples that constructs a proper data manifold and is shown to effectively generate on-manifold samples.
    \item Our proposed VITA achieves state-of-the-art performances on corruption benchmark datasets CIFAR-10-C, CIFAR-100-C, and ImageNet-C. Meanwhile, we demonstrate VITA also boosts adversarial robustness.
\end{itemize}

\section{Related Work}
\label{sec:Related_Work}

\paragraph{Corruption Robustness.}
The human visual system is not easily defrauded by data with various forms of corruption, such as snowflakes, blurring, pixelation, or their combinations.
In contrast, most current deep learning models suffer from severe performance degradation under corrupted data \cite{DBLP:journals/corr/VasiljevicCS16, DBLP:conf/icccn/DodgeK17, DBLP:journals/jmlr/AzulayW19, DBLP:journals/corr/abs-2012-10931}.
For example, \cite{DBLP:conf/nips/GeirhosTRSBW18} reveal that deep neural networks trained on one type of corruption (\textit{e.g.}, salt-and-pepper noise) cannot recognize another unseen type of corruption (\textit{e.g.}, uniform white noise), even though these two kinds of corruptions are indistinguishable to humans.
Currently, research on improving corruption robustness mainly focuses on domain adaptation (\textit{e.g.}, additional operations on the normalization layer) \cite{DBLP:conf/nips/SchneiderRE0BB20, DBLP:journals/corr/abs-2102-02811}, adversarial perturbations \cite{DBLP:conf/iclr/HendrycksD19} and data augmentation \cite{DBLP:conf/iclr/ZhangCDL18, DBLP:conf/iclr/HendrycksMCZGL20, DBLP:journals/corr/abs-2006-16241, DBLP:conf/eccv/KamannR20, DBLP:conf/eccv/RusakSZB0BB20}.
To evaluate the corruption robustness of models,  \cite{DBLP:conf/iclr/HendrycksD19} introduced three comprehensive benchmarks, CIFAR-10-C, CIFAR-100-C and ImageNet-C, for unseen corruption robustness.
Since then, similar datasets on common corruptions have also been proposed in the field of object detection (PASCAL-C, COCO-C and Cityscapes-C) \cite{DBLP:journals/corr/abs-1907-07484} and semantic segmentation \cite{DBLP:conf/eccv/KamannR20}.
These benchmarks demonstrate that the generalization ability of many advanced models under corrupted input still needs to be further improved.

\paragraph{Data Augmentation.}
Data augmentation is one of the most widely studied and effective techniques to improve the corruption robustness of models.
For example, Mixup \cite{DBLP:conf/iclr/ZhangCDL18} is a simple augmentation strategy that performs a linear interpolation between two different classes of samples.
Although it was not specifically proposed to improve corruption robustness, its performance is significantly better than other commonly used data augmentation methods \cite{DBLP:conf/icccn/DodgeK17, DBLP:conf/iccv/YunHCOYC19, DBLP:conf/cvpr/CubukZMVL19}.
Another effective data preprocessing method called AugMix \cite{DBLP:conf/iclr/HendrycksMCZGL20} obtains advanced performance on CIFAR-10-C, CIFAR-100-C and ImageNet-C.
It utilizes a formulation to mix multiple augmented images and adopts a Jensen-Shannon Divergence consistency loss.
Further, \cite{DBLP:conf/eccv/RusakSZB0BB20} demonstrate that data augmented with Gaussian noise can serve as a simple yet very strong baseline for defending against common corruptions.

\section{Proposed Method}
\label{sec:The_Proposed_Method}

\begin{figure*}[!t]
    \centering
    \includegraphics[width=1.0\linewidth]{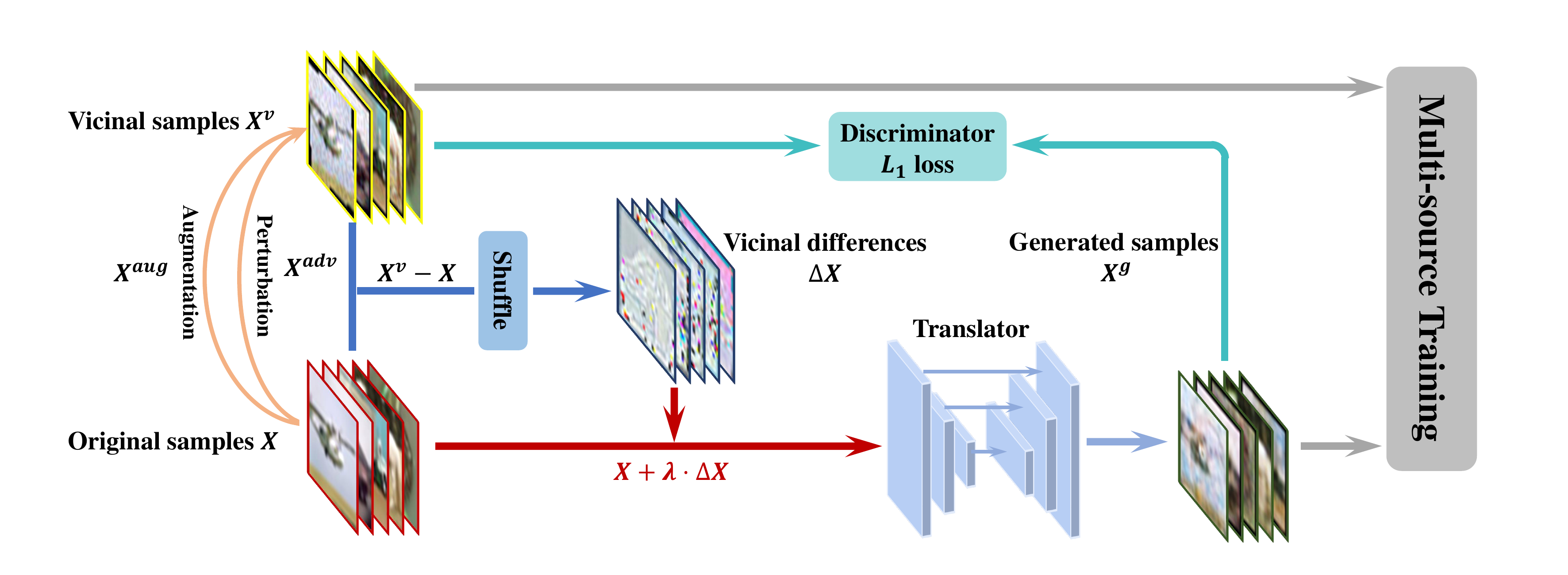}
    \makeatletter\def\@captype{figure}\makeatother\caption{Our proposed VITA includes tangent transfer and multi-source integration.
        We adopt vicinal difference ($\Delta X$) as an approximation of the manifold tangent and use $X + \lambda \cdot \Delta X$ as an initial augmented sample.
        Our generative model includes a $pix2pix$ model, which is designed to generate diverse on-manifold samples.
        The goal of multi-source integration is to learn, based on dataset $\mathcal{D} = \{X^v, Y\}$ ($X^v = \{ X^\text{aug}, X^{adv}\}$), an embedding that imitates the generation process of vicinal samples $P(x^v|x, \delta x)$, where $x$ is an original samples and $\delta x \in \Delta X (\Delta X =  X^v - X$).
        Note that $\delta x$ is a transferred vicinal difference.
        In our robust multi-source training, models are trained with multi-source samples and samples from a well-trained translator.
    }
    \label{fig:framework}
\end{figure*}

In this section, we present the proposed method VITA in detail. 
Our proposed VITA involves two stages.
The first stage, tangent transfer in \ref{subsec:transfer} yields initial augmented samples for improving corruption robustness.
The second stage, multi-source sample integration in \ref{subsec:integration} generates diverse and on-manifold samples to further improve corruption robustness.
In \ref{subsec:robust_training}, we describe how to train with VITA samples. 

\subsection{Tangent Transfer}
\label{subsec:transfer}

\paragraph{Exploiting Shared Manifold Structure. }

\cite{DBLP:conf/nips/BengioM04} used the shared structure of the tangent space of the manifold to mitigate the curse of dimensionality in the previous local manifold learning algorithms.
Generally, in many real-world contexts, there is not just one global manifold but a large number of manifolds that share something about their structure \cite{DBLP:conf/nips/BengioM04, DBLP:conf/cvpr/LasserreBM06}.
However, existing research on data augmentation ignores leveraging this characteristic of the data manifold. 
A simple example is transformations in the image (rotation, lighting, blurring \textit{etc.}).
There is one manifold for each transformation type.
If there are only a few samples of a specific type of transformation during training, it is hard for models to learn a proper data manifold.
As shown in Fig. \ref{fig:benefit_vita}, this leads models to directly memorize these special cases instead of generalizing (\textit{i.e.}, yields the high complexity or unsmooth data manifold).
This, we believe, is the core reason for the model's poor and uneven performance against various forms of corruption.
If the learned structure of the data manifold is shared, models can more accurately characterize a smoother data manifold.

In this work, we use vicinal differences as a rough approximation of manifold tangents.
Transferring vicinal differences enforces the local invariance of the classifier and encourages the classifier to discover shared structures in the tangent planes at different positions.
We argue that approximating the tangent direction of the manifold with the vicinal difference of high-dimensional space can yield more diverse samples, which differs from a model-sensitive measure \cite{DBLP:conf/nips/SimardVLD91, DBLP:conf/nips/SimardLDV96, DBLP:journals/corr/abs-2002-08973}.
Subsequent experiments demonstrated the effectiveness of introducing shuffled vicinal differences.

In specific, vicinal differences are obtained by subtracting the original samples from the vicinal samples.
Here, vicinal samples are crafted through diverse data augmentation operations and adversarial attack methods.
The tangent transfer is realized by adding shuffled vicinal differences.

\paragraph{Transformation Guided by Priors.}
First, we introduce several hand-crafted methods for weakly augmented samples used in our work, including rotation, shearing, translating, cropping and scaling (detailed settings of these augmentations see appendix).
To make the augmented sample more consistent at input level, we need to control the intensity of the change: $||\delta x||_{2} = ||x^\text{aug} - x||_{2} < \epsilon_2$, where $\epsilon$ is a hyperparameter ($\epsilon_2 = 0.5$ by default), $x$ is an original sample and $x^\text{aug}$ is an augmented sample.

\paragraph{Harvesting Adversarial Perturbations.} In contrast to data augmentation, crafting adversarial examples makes use of existing trained models instead of priors from human experience.
Thus, we employ several algorithms to generate adversarial examples, including the fast gradient sign method (FGSM) \cite{DBLP:journals/corr/GoodfellowSS14}, projected gradient descent (PGD) \cite{DBLP:conf/iclr/MadryMSTV18}, momentum iterative method \cite{DBLP:conf/cvpr/DongLPS0HL18}, C\&W method \cite{DBLP:conf/sp/Carlini017} and Elastic-Net Attack \cite{DBLP:conf/aaai/ChenSZYH18} (detailed hyper-parameter settings see appendix).
For different well-trained models, the adversarial examples of a given seed $x$ will vary because of the diverse loss functions and hypothetical spaces.
However, they all reflect the local structure of the seed $x$ from the different perspectives.
For adversarial examples, we also control the magnitude of perturbations.
For all datasets, we use the $\ell_\infty-$ and $\ell_2-$ adversarial attack methods with fixed budget of $\epsilon_\infty = 0.031$ and $\epsilon_2 = 0.5$.

\subsection{Multi-Source Samples Integration}
\label{subsec:integration}

\paragraph{Utilizing Differences via a Generative Model}

In the following experiments, we integrate all the crafted samples, which include the multi-source vicinal examples from data augmentation and adversarial perturbations, into a dataset $\mathcal{D} = \{X^\text{aug},X^\text{adv}, Y\}$ and treat them equivalently. Note that $x^v \in \{ X^\text{aug}, X^{adv}\}$ is the input data and $y \in Y$ is the label of $x^v$.
Our goal is to learn, based on dataset $\mathcal{D}$, an embedding that imitates the generation process of vicinal samples $P(x^v|x, \delta x)$ given $x$ and $\delta x$.
Note that $\delta x \in \Delta X$ is a transferred vicinal difference.
To this end, we use an adversarial loss \cite{DBLP:conf/nips/GoodfellowPMXWOCB14} to implement the embedding and build two basic models: a sample-to-sample translation model $T$ and a discriminative model $D$.

Mapping from input distribution to output distribution in high-dimensional space is challenging \cite{DBLP:conf/nips/ZhuZPDEWS17}.
Thus, we start with the pix2pix framework \cite{DBLP:conf/cvpr/IsolaZZE17}, which has previously been shown to produce high-quality results for various image-to-image translation tasks.
In our appendix, we also elaborate on the ablation experiments related to the translator, discriminator and the impact of using a more complex framework such as BicycleGAN \cite{DBLP:conf/nips/ZhuZPDEWS17}.

\paragraph{Translator.}
The sample with vicinal difference may not fall on the local data manifold because of the complex high-dimensional data space.
Thus, we need to build a translator to embed the raw intermediate product $x + \delta x$ into the on-manifold vicinal sample $x^g = T(x + \delta x)$.
As shown in Fig. \ref{fig:framework}, inputs to our translator are the samples added with shuffled vicinal differences, and outputs are desired samples on the data manifold.
By default, when training the translator, we include the same proportion of augmented and adversarial data.
Our translator takes advantage of a U-Net structure \cite{DBLP:conf/miccai/RonnebergerFB15}, which enables the transmission of hierarchical information across a network by skipping layer connections.
In our experiments, the U-Net structure has shown its effectiveness to preserve the vicinal differences.
For more details on the superiority of the U-Net architecture, see our ablation experiments in the appendix.

\paragraph{Discriminator.}
We denote our discriminator as $D$, and $D(x^g)$ indicates the probability that a generated sample $x^g$ comes from the real vicinity.
In our network, we use a $1 \times 1$ PatchGAN \cite{DBLP:conf/cvpr/IsolaZZE17} discriminator by default.
PatchGAN discriminator is a type of discriminator for generative adversarial networks which only penalizes structures at the scale of the local image patches.
The PatchGAN discriminator tries to classify whether each patch in an image is real or fake.

\paragraph{Objective Function.} The embedding of the translator and discriminator is configured using neural nets.
We describe the main part of loss below.
\begin{equation}
    \begin{split}
        \mathcal{L}_{\text{GAN}}(\T,\D) & = \mathds{E}_{\A,\B\sim p(\A,\B)}[\log(\D(\A,\B))] + \\
        & \mathds{E}_{\A\sim p(\A),\z\sim p(\z)}[ \log(1-\D(\A,\T(\A + \z)))]
    \end{split}
    \label{eqn:Lgan}
\end{equation}
Here, $p(\A, \B)$ represents the joint distribution of the original samples distribution $x$ and the vicinal samples $x^v$ distribution, $p(\A)$ denotes the distribution of $x$, and $p(\z)$ represents the the distribution of vicinal differences.

\subsection{Robust Training Process}
\label{subsec:robust_training}

We argue that multi-source samples can provide more vicinal information, as shown in Fig. \ref{fig:benefit_vita}.
% For example, over-sampling strong adversarial examples will cause the model to overfit, while effectively adopting of the shared local information provided by other samples can lead the model not to over focus on a certain type of perturbations or augmentation.
In order to make full use of multi-source samples, we divide samples into three categories according to their source, namely the weakly augmented samples, the samples added with shuffled adversarial perturbations, the samples generated via VITA.
During training, we split the samples in each batch into weakly augmented samples (25\%), shuffled perturbations samples (25\%) and generated samples via VITA (50\%).
Among the generated samples, half of the vicinal differences for the translator come from weakly augmented samples, and half of them are generated from shuffled adversarial perturbations.
More specifically, we apply a Jensen-Shannon Divergence consistency loss as a regularization term to enforce a consistent embedding by the classifier across further diverse augmentation.
More details on our robust training process can be found in our supplementary materials.

\section{Experiments}
\label{sec:Experiment}

\paragraph{Dataset.}
CIFAR-10 (10 categories) and CIFAR-100 (100 categories) both contain small $32 \times 32 \times 3$ colour images, with 50k for training and 10k for testing.
The ImageNet \cite{DBLP:conf/cvpr/DengDSLL009} dataset includes 1,000 classes and contains approximately 1.2 million images annotated according to the WordNet hierarchy.
To evaluate the corruption robustness of models, we conduct experiments on the CIFAR-10-C, CIFAR-100-C and ImageNet-C datasets \cite{DBLP:conf/iclr/HendrycksD19}.
These datasets are constructed by corrupting the original images from the CIFAR-10, CIFAR-100 and ImageNet test sets.
Specifically, the CIFAR-10-C, CIFAR-100-C, and ImageNet-C datasets consist of 15 types of algorithmically generated corruptions from noise, blur, weather, and digital categories.
Each type of corruption has five levels of severity, resulting in 75 distinct corruptions.
Since these datasets are used to measure model performance under data shifts, we take care not to introduce the 15 corruptions into the VITA process and robust training procedure.

\begin{figure}[!t]
    \centering
    \includegraphics[width=\linewidth]{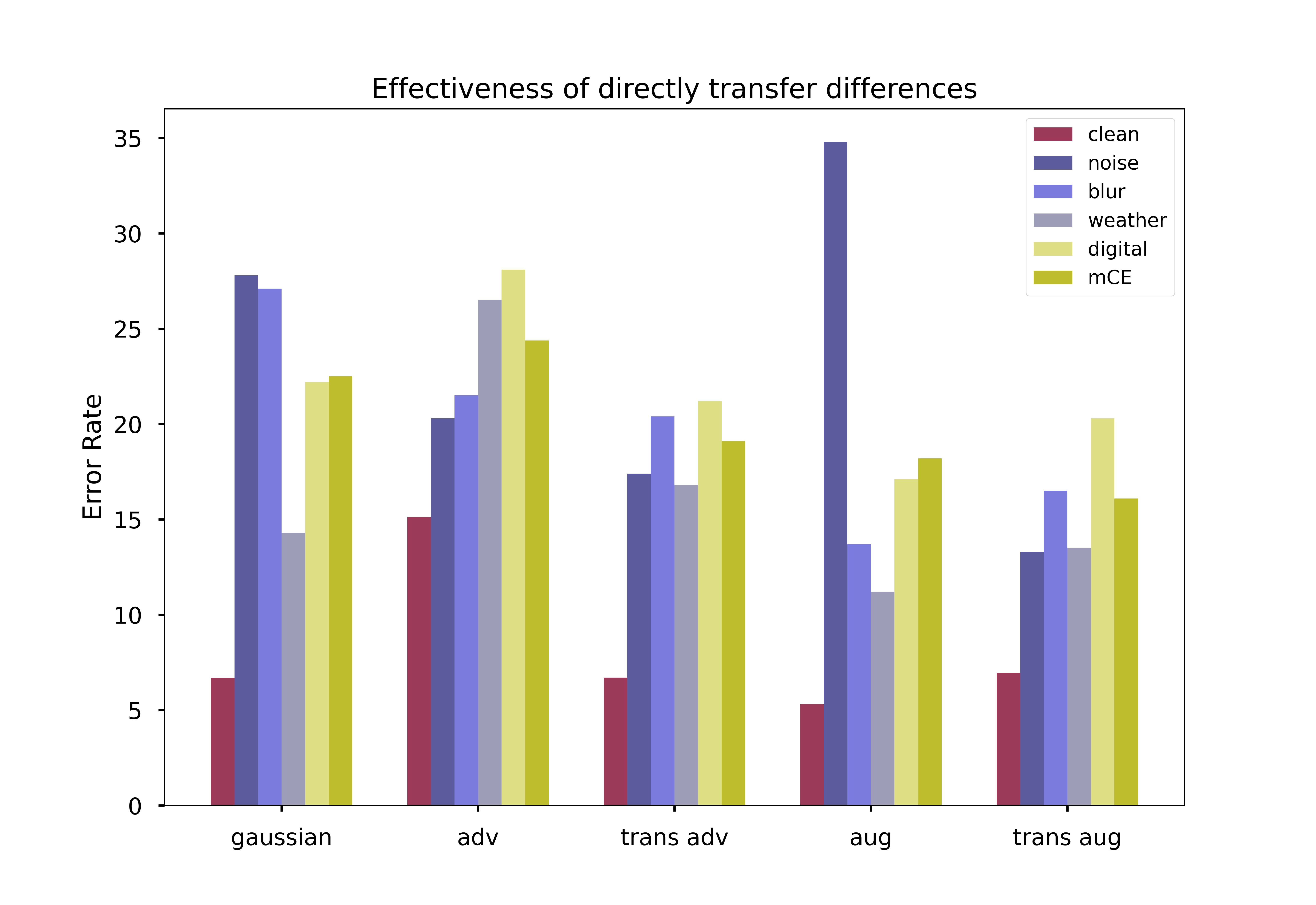}
    \makeatletter\def\@captype{figure}\makeatother
    \caption{Effectiveness of directly transferring differences.
        We compare the corruption robustness of models trained with different augmentation methods, including Gaussian noise (Gaussian), adversarial perturbations (adv), transferred perturbations (trans adv), weak augmentation (aug), transferred differences from weak augmentation (trans aug).
        We evaluate corruption robustness (error rate, the lower, the better) on CIFAR-10-C with an AllConvNet architecture.
        Here, \textit{clean} indicates the performance on clean (uncorrupted) images, \textit{noise/blur/weather/digital} is robustness towards the corresponding corruption types and \textit{mCE} is mean corruption error for all types of corruption.
        As seen, transferring vicinal differences improves robustness significantly.}
    \label{fig:direct_transfer}
\end{figure}

\begin{table*}[!t]
    \setlength\tabcolsep{3pt}
    \centering
    \normalsize
    \caption{Mean corruption error (mCE) on CIFAR-10-C and CIFAR-100-C. Average classification error as percentages. Across several architectures, our method obtains obvious enhancement in corruption robustness. Specifically, we reduced the error rate of corrupted data to 8.9 on ResNeXt.}
    \begin{tabular}{ ll|ccccccccc }
        \hline
                                  &            & Standard & Cutout & Mixup  & CutMix & AutoAug & Adv Train & AugMix & ME-ADA        & VITA
        \\
        \hline
        \parbox[t]{20mm}{\multirow{4}{*}{\rotatebox{0}{CIFAR-10-C}}}
                                  & AllConvNet & 30.8     & 32.9   & 24.6   & 31.3   & 29.2    & 28.1      & 15.0   & 21.8          & \textbf{10.6} \\
                                  & DenseNet   & 30.7     & 32.1   & 24.6   & 33.5   & 26.6    & 27.6      & 12.7   & 23.1          & \textbf{9.7}  \\
                                  & WideResNet & 26.9     & 26.8   & 22.3   & 27.1   & 23.9    & 26.2      & 11.2   & 16.7          & \textbf{9.5}  \\
                                  & ResNeXt    & 27.5     & 28.9   & 22.6   & 29.5   & 24.2    & 27.0      & 10.9   & 16.6          & \textbf{8.9}  \\
        \hline
        \multicolumn{2}{c|}{Mean} & {29.0}     & {30.2}   & {23.5} & {30.3} & {26.0} & {27.2}  & {12.5}    & {19.5} & \textbf{9.7}                  \\
        \Xhline{3\arrayrulewidth}
        \parbox[t]{20mm}{\multirow{4}{*}{\rotatebox{0}{CIFAR-100-C}}}
                                  & AllConvNet & 56.4     & 56.8   & 53.4   & 56.0   & 55.1    & 56.0      & 42.7   & 48.8          & \textbf{36.3} \\
                                  & DenseNet   & 59.3     & 59.6   & 55.4   & 59.2   & 53.9    & 55.2      & 39.6   & 52.2          & \textbf{35.4} \\
                                  & WideResNet & 53.3     & 53.5   & 50.4   & 52.9   & 49.6    & 55.1      & 35.9   & 47.2          & \textbf{34.4} \\
                                  & ResNeXt    & 53.4     & 54.6   & 51.4   & 54.1   & 51.3    & 54.4      & 34.9   & 42.7          & \textbf{31.5} \\
        \hline
        \multicolumn{2}{c|}{Mean} & {55.6}     & {56.1}   & {52.6} & {55.5} & {52.5} & {55.2}  & {38.3}    & {47.7} & \textbf{34.4}                 \\
        \Xhline{3\arrayrulewidth}
    \end{tabular}
    \label{tab:cifar}
    % \vspace{-7.5pt}
\end{table*}

\paragraph{Metric.}
The clean error is the usual classification error on uncorrupted test images.
In terms of measuring corruption robustness, we use mean error at five different intensities or levels of severity, i.e. $1 \leq s \leq 5$.
Let $E_{c,s}$ denote the test error of corrupted images from corruption type $c$ and under severity level $s$.
For CIFAR datasets, we use the mean corruption error ($mCE$) over fifteen corruptions and five severities, i.e. $mCE = 1 / 75 \sum^{15}_{c=1} \sum^5_{s=1} E_{c,s}$.
For ImageNet, we follow the convention of normalizing the corruption error by the corruption error of AlexNet \cite{DBLP:conf/nips/KrizhevskySH12}, i.e. $ CE_c = \sum^5_{s=1} E_{c,s} / \sum^5_{s=1} E^{AlexNet}_{c,s} $.
The mean of the 15 corruption errors gives us the $mCE = 1 / 15 \sum^{15}_{c=1} CE_c$.

\subsection{Effectiveness of Transferring Differences}

\paragraph{Setup.}
Our verification experiment is based on the AllConv \cite{DBLP:journals/corr/SpringenbergDBR14} network, trained on the clean CIFAR-10 dataset and evaluated on the CIFAR-10-C dataset.
We mainly compare the impact of five different inputs on the corruption robustness.
These inputs including samples added with Gaussian noise (\textit{Gaussian}, mean value 0, standard deviation 0.5), adversarial perturbations (\textit{adv}), transferred perturbations (\textit{trans adv}).
Training with weakly augmented samples (\textit{aug}), and samples with transferred differences from weakly augmented samples (\textit{trans aug}) are also included.
In our supplementary materials, we conduct an ablation study on the types of augmentation and perturbations. And it further reveals the effectiveness of transferring difference on alleviating the model's sensitivity to certain types of local direction.

\paragraph{Results.}
Our first experiment verifies the validity of transferring vicinal differences.
Our main findings are as follows. The first and most important thing is that transferring vicinal differences can improve the corruption robustness of, as shown in Fig. \ref{fig:direct_transfer}.
In particular, transferring vicinal differences from weakly augmented samples can greatly improve robustness toward common corruptions.
Also, the improvement brought by transferring vicinal differences is significantly greater than the addition of Gaussian noise.
Interestingly, we discover that transferring adversarial perturbations improves robustness to noise corruption more effectively than adding Gaussian noise.

\subsection{Effectiveness of VITA}

In this part, we verify the effectiveness of our vicinal information fusion framework, that is, adding samples generated by VITA to the multi-source robust training process (as described in \ref{subsec:robust_training}).

\begin{table*}[!t]
    \small
    \centering
        \caption{Clean error, mean corruption error (mCE) and all types of corruption error rate values for various methods on ImageNet-C.
            We compare against other data augmentation methods for improving the corruption robustness. Our proposed VITA hugely improves corruption robustness and achieves balanced performance toward different corrupted images.}
        {\setlength\tabcolsep{1.5pt}%
            \begin{tabular}{@{}l  c | c c c | c c c c | c c c  c | c c c c@{} | c }
                \multicolumn{2}{c}{} & \multicolumn{3}{c}{Noise}      & \multicolumn{4}{c}{Blur} & \multicolumn{4}{c}{Weather} & \multicolumn{4}{c}{Digital} & \multicolumn{1}{c}{}                                                                                                                                                                                  \\
                \cline{1-18}
                Network              & \multicolumn{1}{c|}{\,Clean\,} & Gauss.
                                     & Shot                           & Impulse                  & Defocus                     & Glass                       & Motion               & Zoom        & Snow        & Frost       & Fog         & Bright      & Contrast    & Elastic     & Pixel       & JPEG        & {\,\textbf{mCE}\,}                               \\ \hline
                Standard             & 23.9                           & 79                       & 80                          & 82                          & 82                   & 90          & 84          & 80          & 86          & 81          & 75          & 65          & 79          & 91          & 77                 & 80          & 80.6          \\
                Patch Uniform        & 24.5                           & 67                       & 68                          & 70                          & 74                   & 83          & 81          & 77          & 80          & 74          & 75          & 62          & 77          & 84          & 71                 & 71          & 74.3          \\
                AutoAug              & 22.8                           & 69                       & 68                          & 72                          & 77                   & 83          & 80          & 81          & 79          & 75          & 64          & 56          & 70          & 88          & 57                 & 71          & 72.7          \\
                % Random AA*                              & 23.6                           & 70                       & 71                          & 72                          & 80                   & 86          & 82          & 81          & 81          & 77          & 72          & 61          & 75          & 88          & 73                 & 72          & 76.1          \\
                MaxBlur pool         & 23.0                           & 73                       & 74                          & 76                          & 74                   & 86          & 78          & 77          & 77          & 72          & 63          & 56          & 68          & 86          & 71                 & 71          & 73.4          \\
                SIN                  & 27.2                           & 69                       & 70                          & 70                          & 77                   & 84          & 76          & 82          & 74          & 75          & 69          & 65          & 69          & 80          & 64                 & 77          & 73.3          \\
                AugMix               & \textbf{22.4}                  & 65                       & 66                          & 67                          & 70                   & 80          & 66          & 66          & 75          & 72          & 67          & 58          & 58          & 79          & 69                 & 69          & 68.4          \\
                % AugMix+SIN           & 25.2                           & 61                       & 62                          & 61                          & 69                   & 77   & 63   & 72    & 66  & 68     & 63       & 59      & 52    & 74   & 60                 & 67 & 64.9 \\
                DeepAug              & 23.3                           & 49                       & 50                          & 47                          & 59                   & 73          & 65          & 76          & 64          & 60          & 58          & 51          & 61          & 76          & 48                 & 67          & 60.4          \\
                % DeepAug+AugMix       & 24.2                           & 46                       & 45                          & 44                          & 50                   & 64   & 50   & 61    & 58  & 57     & 54       & 52      & 48    & 71   & 43                 & 61 & 53.6 \\
                ANT                  & 23.9                           & \textbf{39}              & \textbf{40}                 & \textbf{39}                 & 68                   & 78          & 73          & 77          & 71          & 66          & 68          & 55          & 69          & 79          & 63                 & 64          & 63.3          \\
                % ANT+SIN              & 25.9                           & 43                       & 44                          & 43                          & 71                   & 74   & 69   & 79    & 60  & 58     & 55       & 56      & 59    & 73   & 57                 & 67 & 60.5 \\
                \cline{1-18}

                VITA                 & 25.4                           & 40                       & 41                          & 41                          & \textbf{47}          & \textbf{61} & \textbf{51} & \textbf{59} & \textbf{57} & \textbf{58} & \textbf{55} & \textbf{49} & \textbf{47} & \textbf{69} & \textbf{44}        & \textbf{62} & \textbf{52.1} \\
                % VITA+SIN             & 25.9                           & 43                       & 44                          & 43                          & 71                   & 74   & 69   & 79    & 60  & 58     & 55       & 56      & 59    & 73   & 57                 & 67 & 60.5
                %
            \end{tabular}}

        \label{tab:imagenet-table}
    
\end{table*}

\begin{figure}[!t]
    \centering
    \includegraphics[width=\linewidth]{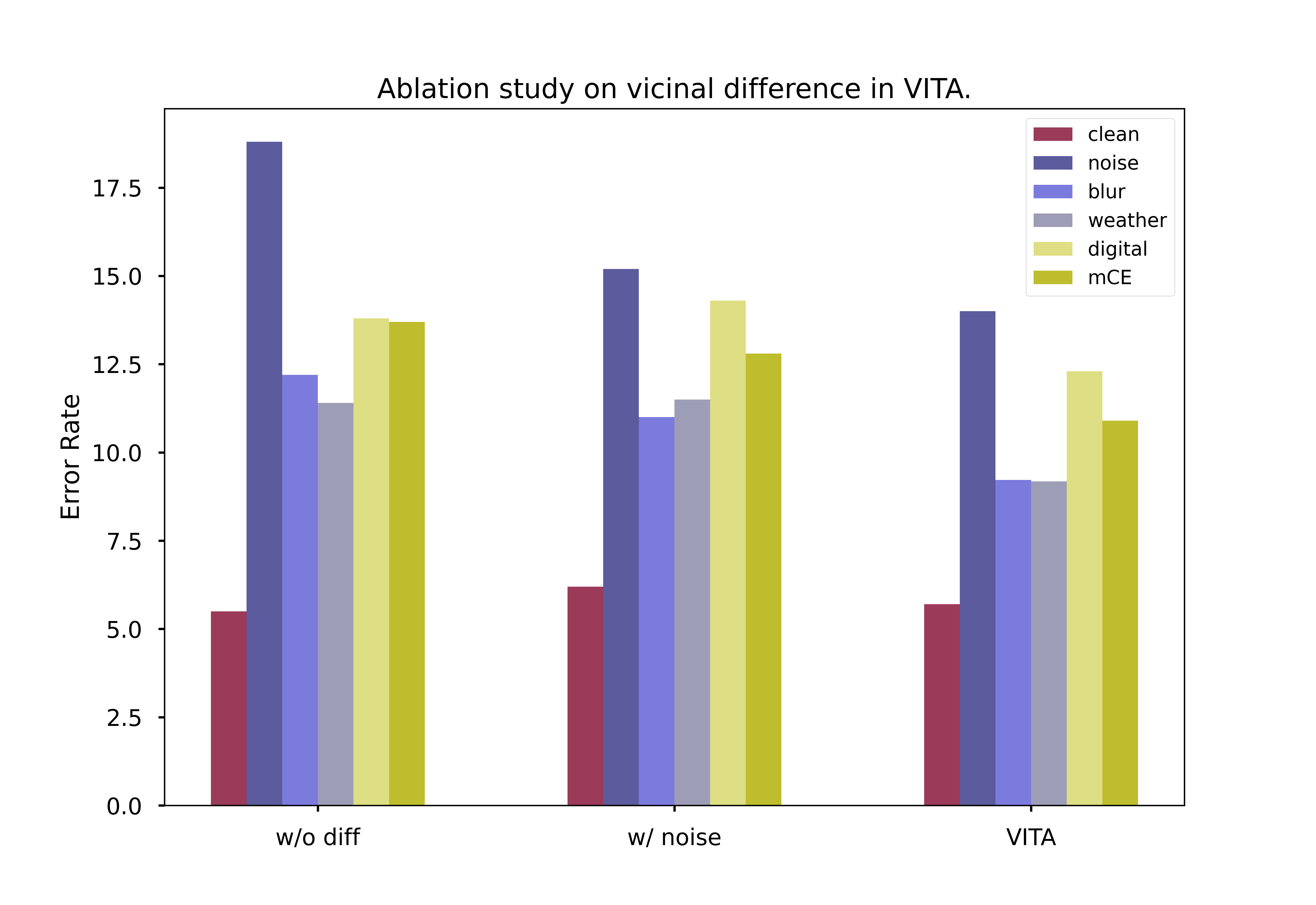}
    \makeatletter\def\@captype{figure}\makeatother
    \caption{Ablation study on vicinal differences in VITA, demonstrating the necessity of transferred vicinal differences as input to a translator.
        We evaluate corruption robustness (error rate, the lower, the better) on CIFAR-10-C with an AllConvNet architecture.
        Here, \textit{w/o diff} is the translator trained and inferred without transferred differences, and \textit{w/ noise} is the translator trained and inferred with the addition of Gaussian noise.
        Training process is the same setting as for default robust training process (50\% gen + 25 \% adv + 25\% aug).
        As seen, a translator trained with Gaussian noise or without vicinal differences (\textit{i.e.} merely original data) performs worse than VITA.    }
    \label{fig:difference_ablation}
\end{figure}

\paragraph{CIFAR Training Settings.} In the following experiments, we choose the same network architectures as AugMix \cite{DBLP:conf/iclr/HendrycksMCZGL20}, including All Convolutional Network \cite{DBLP:journals/corr/SpringenbergDBR14}, DenseNet-BC ($k=2,d=100$) \cite{DBLP:conf/cvpr/HuangLMW17}, 40-2 WideResNet \cite{DBLP:conf/bmvc/ZagoruykoK16} and ResNeXt-29 ($32 \times 4$) \cite{DBLP:conf/cvpr/XieGDTH17}. We use stochastic gradient descent with an initial learning rate of 0.1 and \textit{ReduceOnPlateau} scheduler. We train all architectures over 150 epochs.

\paragraph{CIFAR Results.}
We perform a comprehensive evaluation to compare with a total of 7 advanced augmentation methods, including Cutout \cite{DBLP:journals/corr/abs-1708-04552}, Mixup \cite{DBLP:conf/iclr/ZhangCDL18}, CutMix \cite{DBLP:conf/iccv/YunHCOYC19}, AutoAug \cite{DBLP:conf/cvpr/CubukZMVL19}, adversarial training (Adv Train) \cite{DBLP:conf/sp/Carlini017}, AugMix \cite{DBLP:conf/iclr/HendrycksMCZGL20}, ME-ADA \cite{DBLP:conf/nips/000300M20}.
Compared to the standard data augmentation baseline (mean of four different architectures), our approach achieves 19.3\% lower $mCE$ as shown in Fig. \ref{tab:cifar}.
Compared to AugMix, which is the current state of the art on CIFAR-10-C and CIFAR-100-C, our method obtains significant performance improvement under various network architectures.
Specifically, we achieve a 4.4\% (CIFAR-10-C) and 6.4\% (CIFAR-100-C) performance improvement in mCE under the AllConvNet compared with AugMix.

\begin{table*}[!t]
   % \small 
   \centering
   \caption{Evaluations (test accuracy) of deep models (WRN-34-10) on the CIFAR-10 dataset. Results of TRADES ($\beta = 1.0$ and $6.0$) are reported in~\cite{DBLP:conf/icml/ZhangYJXGJ19}. Results of FAT for TRADES are reported in~\cite{DBLP:conf/icml/ZhangXH0CSK20}. Our proposed framework of VITA and multi-source integrated training can also improve the adversarial robustness of the model.}
   \label{tab:sota_result_trades}
   % \resizebox{\textwidth}{12mm}{
   \begin{tabular*}{\hsize}{@{}@{\extracolsep{\fill}}clcccccc@{}}
      % \begin{tabular}{clcccc}
      \hline
      \multicolumn{2}{c|}{Defense} & \multicolumn{1}{c}{Natural} & \multicolumn{1}{c}{FGSM} & \multicolumn{1}{c}{PGD-20} & \multicolumn{1}{c}{C$\&$W$_{\infty}$} & \multicolumn{1}{c}{PGD-100} & \multicolumn{1}{c}{AutoAttack} \\
      \hline
      \multicolumn{2}{c|}{TRADES ($\beta = 1.0$)} & 88.64 & 56.38 & 49.14 & - & - & -\\
      \multicolumn{2}{c|}{FAT for TRADES} & \textbf{89.94} & 61.00 & 49.70 & 49.35 & 48.35 & 47.22 \\
      \multicolumn{2}{c|}{VITA for Adv. Training} & 89.35  & \textbf{68.02}  & \textbf{52.33} & \textbf{50.21} & \textbf{50.04} & \textbf{48.38} \\
      \hline
      \multicolumn{2}{c|}{TRADES ($\beta = 6.0$)} & 84.92 & 61.06 & 56.61 & 54.47 & 55.47 & 53.08 \\
      \multicolumn{2}{c|}{FAT for TRADES} & \textbf{86.60} & 61.97 & 55.98 & 54.29 & 55.34 & 53.27 \\
      \multicolumn{2}{c|}{VITA for Adv. Training} & 85.75 & \textbf{67.99} & \textbf{57.63} & \textbf{55.32} & \textbf{56.87} & \textbf{54.35} \\
      \hline
   \end{tabular*}
   %\vskip -0.1in
   \vskip1ex%
   \vskip -0.1in
\end{table*}

\paragraph{ImageNet Training Settings.}
We use ResNet-50 as the backbone of our model trained on ImageNet. The training scheme follows AugMix; that is, we apply a small learning rate for the first five epochs to warm up the training and then apply a decayed learning rate for the remaining epochs.
In addition to AugMix with standard training, we also compare our method with DeepAug \cite{DBLP:journals/corr/abs-2006-16241}, stylized image training (SIN) \cite{DBLP:conf/iclr/GeirhosRMBWB19} and adversarial noise training (ANT) \cite{DBLP:journals/corr/abs-1907-07484}.
Stylized image training refers to the method in which the model is not only trained on the original dataset but also on stylized ImageNet samples \cite{DBLP:conf/iclr/GeirhosRMBWB19}.

\paragraph{ImageNet Results.}
We perform a large-scale evaluation to compare with a total of 7 advanced augmentation methods, including Patch Uniform \cite{DBLP:journals/corr/abs-1906-02611}, AutoAug \cite{DBLP:conf/cvpr/CubukZMVL19}, MaxBlur pool \cite{DBLP:conf/icml/Zhang19}, SIN \cite{DBLP:conf/iclr/GeirhosRMBWB19}, AugMix \cite{DBLP:conf/iclr/HendrycksMCZGL20}, DeepAug \cite{DBLP:journals/corr/abs-2006-16241}, ANT \cite{DBLP:conf/eccv/RusakSZB0BB20}.
Although our method has a slight drop in accuracy on clean samples, it achieves 52.1\% mCE as shown in Fig. \ref{tab:imagenet-table}, compared to AugMix with 68.4\% and ANT with 63.3\%.
Moreover, even without Gaussian noise in the training phase, our method's performance in terms of noise corruption is close to ANT's (adversarial noise training).
This shows the ability of VITA to promote balanced performance on different corruption types.

\begin{figure}[!t]
    \centering
    \includegraphics[width=\linewidth]{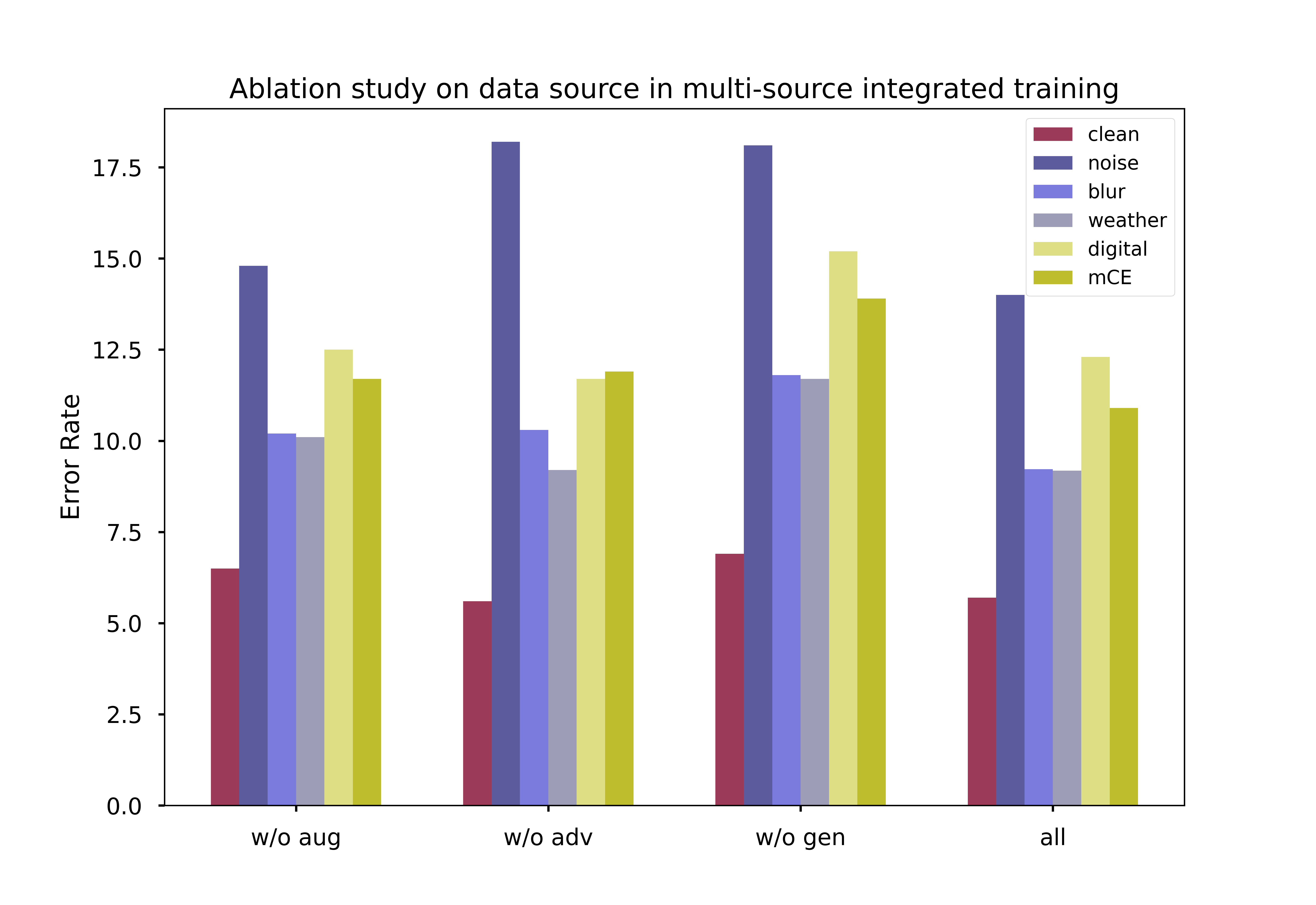}
    \makeatletter\def\@captype{figure}\makeatother
    \caption{Ablation study on the data source in multi-source robust training reveals the importance of training with samples from VITA.
        We evaluate corruption robustness (error rate, the lower the better) on CIFAR-10-C with an AllConvNet architecture.
        Here, \textit{w/o aug/adv/gen} indicate that specific types of samples lack in our multi-source robust training.
        For example, we train with 50\% of samples with shuffled perturbations and 50\% samples generated by VITA during training under the \textit{w/o aug} setting.
        Obviously, when robust training does not include samples generated by VITA, corruption robustness decreases significantly.}
    \label{fig:source_ablation}
\end{figure}

\subsection{Ablation Study on Vicinal Differences in VITA}

\paragraph{Setup.}
% In VITA, we train the translator with transferred vicinal differences.
In this part, we investigate the effects of inputs to the translator on the quality of the generated sample (i.e. the impact on improving the robustness of corruption).
We conduct our ablation study on CIFAR-10-C with an AllConvNet architecture.
We add two sets of control experiments: one with the translator trained and inferred without vicinal differences (\textit{w/o diff}) and one with Gaussian noise (\textit{w/ noise}).

\paragraph{Results.}
As illustrated in Fig. \ref{fig:difference_ablation}, robust training with samples generated by the translator trained without vicinal differences is less robust against corruption, particularly noise corruption.
Although training with samples from a noise-added translator improves performance against noise corruption, it remains significantly poorer than VITA on other types of corruption.

\subsection{Ablation Study on Data Source in Training}

\paragraph{Setup.}
In our multi-source robust training, we have three types of samples (${X^{aug}, X^{adv}, X^{gen}}$).
This part aims to determine which data source or combination of data sources most significantly contributes to the enhancement of corruption robustness and performance on a clean test set (for more ablation experiments on data sources and dataset size, see appendix).
In Fig. \ref{fig:source_ablation}, \textit{w/o aug} (\textit{w/o adv/gen} are similar) indicates that we only remove augmented samples during training (i.e. 50\% \textit{adv} + 50\% \textit{gen}).
It should be emphasized that our data source ablation experiment is conducted during the multi-source robust training stage, and we still use multi-source samples when training the translator.

\paragraph{Results.}
Firstly, as can be seen from Fig. \ref{fig:source_ablation}, the samples generated by VITA play an important role in improving the corruption robustness of the model.
Without samples generated from VITA (i.e. \textit{w/o gen}), the model performs poorly on various types of corruption.
When the model is trained with samples generated from VITA and one type of vicinal samples (i.e. \textit{w/o aug} or \textit{w/o adv}), its corruption robustness is better than when training with both types of vicinal samples (i.e. \textit{w/o gen}).
However, its corruption robustness is still not as good as when vicinal samples are integrated from all sources.
Besides, we can find that the combination of samples with shuffled adversarial perturbations and samples generated by VITA performs best (compared with \textit{w/o adv} and \textit{w/o gen}).
This shows the superiority of the weakly augmented samples we generated compared to the common weakly augmented samples.

\subsection{Robustness Towards Adversarial Attack}

\paragraph{Adversarial Training Settings.}
Recent works \cite{DBLP:conf/eccv/RusakSZB0BB20, DBLP:conf/icml/GilmerFCC19} substantiate the claim that increased robustness against regular or universal adversarial perturbations \cite{DBLP:conf/sp/Carlini017, DBLP:conf/cvpr/Moosavi-Dezfooli17} does not imply increased robustness against common corruptions.
In this part, we discuss whether our generated samples from VITA can be integrated into the existing adversarial training process to improve the adversarial robustness of the model.
All images of CIFAR-10 are normalized into [0, 1].
The adversarial test data are bounded by $l_{\infty}$ perturbations with $\epsilon_{test} = 0.031$.
We use the same settings as the corruption robustness evaluation experiment to train our image-to-image translation framework.
The only difference is the regularization terms in multi-source robust training.
Here, we use strong adversarial examples via the regularization term proposed by TRADES \cite{DBLP:conf/icml/ZhangYJXGJ19}.
Using a regularization term from TRADES \cite{DBLP:conf/icml/ZhangYJXGJ19} and an early stopping scheme from FAT \cite{DBLP:conf/icml/ZhangXH0CSK20}, we deploy multi-source adversarial training to verify the effectiveness of samples generated by VITA towards adversarial robustness.
The backbone of our network is Wide ResNet. Models are trained using SGD with 0.9 momenta for 100 epochs, with the initial learning rate of 0.01 divided by ten at epoch 60.
Our supplementary materials contain further information about training and evaluating with additional models.

\paragraph{Adversarial Robustness Results.}
In Table \ref{tab:sota_result_trades}, we can see that injecting the data generated by our method clearly improves the model's adversarial robustness against various adversarial attack methods.
The usefulness of our suggested framework in strengthening model robustness against FGSM attack is particularly obvious.
Although our VITA is not specifically designed for defending various adversarial attacks, we can easily increase adversarial robustness by simply altering the regularization terms.

\section{Conclusion}
\label{sec:Conclusion}
In this work, we propose a multi-source vicinal transfer augmentation (VITA) method to mitigate performance degradation caused by off-manifold samples.
The proposed VITA consists of two components: tangent transfer and integration of multi-source vicinal samples.
To the best of our knowledge, our work is the first to reveal the effectiveness of tangents transfer for improving corruption robustness.
Experimental results show that our proposed VITA obtains state-of-the-art performance on the image corruption benchmarks (CIFAR-10-C, CIFAR-100-C and ImageNet-C).
% We show that VITA can boost adversarial robustness as well.

\clearpage
\linespread{1}\selectfont
\section*{Acknowledgment}
This work is supported by the National Natural Science Foundation of China under Grant No. 61972188 and No. 62122035 and the National
Key R\&D Program of China under Grant No. 2021ZD0111700.

\bibliography{aaai22}

\clearpage
% Use conservative float limits for the large appendix charts.
\renewcommand{\topfraction}{0.7}
\renewcommand{\dbltopfraction}{0.7}
\renewcommand{\textfraction}{0.2}
\renewcommand{\floatpagefraction}{0.5}
\renewcommand{\dblfloatpagefraction}{0.5}
\setcounter{topnumber}{2}
\setcounter{dbltopnumber}{2}
\setcounter{totalnumber}{3}
% Align float-only appendix pages at the top with consistent spacing.
\makeatletter
\setlength{\@fptop}{0pt}
\setlength{\@fpsep}{12pt}
\setlength{\@fpbot}{0pt plus 1fil}
\setlength{\@dblfptop}{0pt}
\setlength{\@dblfpsep}{12pt}
\setlength{\@dblfpbot}{0pt plus 1fil}
\makeatother
\appendix
% appendix

\section{Corruption Robustness Benchmarks}

\begin{algorithm}[t]
    \caption{Pseudocode of multi-source robust training in a PyTorch-like style.}
    \label{alg:code}
    \algcomment{\fontsize{7.2pt}{0em}\selectfont \texttt{aug\_ops}: pre-defined augmentations; \texttt{adv\_ops}: pre-defined adversarial attack; \texttt{pgd\_attack}: PGD adversarial attack.
        %\vspace{-1.em}
    }
    \definecolor{codeblue}{rgb}{0.25,0.5,0.5}
    \lstset{
        backgroundcolor=\color{white},
        basicstyle=\fontsize{7.2pt}{7.2pt}\ttfamily\selectfont,
        columns=fullflexible,
        breaklines=true,
        captionpos=b,
        commentstyle=\fontsize{7.2pt}{7.2pt}\color{codeblue},
        keywordstyle=\fontsize{7.2pt}{7.2pt},
        %  frame=tb,
    }
    \begin{lstlisting}[language=python]
# lambda: hyper-paragraph for diff. addition
# netT: well-trained translator
# train_mode: train for adv. or corruption robustness
# model: target model for training
# beta: hyper-parameter for regularization

for x, y in loader:
   x1, x2 = torch.chunk(x, 2, dim=0) # halves x
   x_aug = aug_ops(x1) # augmentation
   x_adv = adv_ops(x2) # adversarial attack

   # shuffle vicinal differences
   aug_diff = shuffle(x_aug - x1)
   adv_diff = shuffle(x_adv - x2)

   # (optional) train with shuffled diff
   x_aug = x1 + aug_diff
   x_adv = x2 + adv_diff

   # generate samples via translator
   x_vita_aug = netT(x1 + lambda * aug_diff)
   x_vita_adv = netT(x2 + lambda * adv_diff)

   # concatenate multi-source samples
   x_all = torch.cat([x_aug, x_adv, x_vita_aug, x_vita_adv], dim=0)

   # adjust reg. term when training for adv. robustness
   if train_mode == 'adversarial_training':
      x_reg = pgd_attack(x_all, model)
   else:
      x_reg = aug_ops(x_all) # for corruption robustness

   # compute logits
   logits_all = model(x_all)
   logits_reg = model(x_reg)

   # concatenate labels
   y_all = torch.cat([y, y], dim=0)

   # loss function
   loss = CrossEntropyLoss(logits_all, y_all)
   loss += beta * KLDivLoss(log_softmax(logits_all), log_softmax(logits_reg))

   # SGD update: model
   loss.backward()
   
\end{lstlisting}
\end{algorithm}

\begin{figure*}
   \centering
   \includegraphics[scale=0.34]{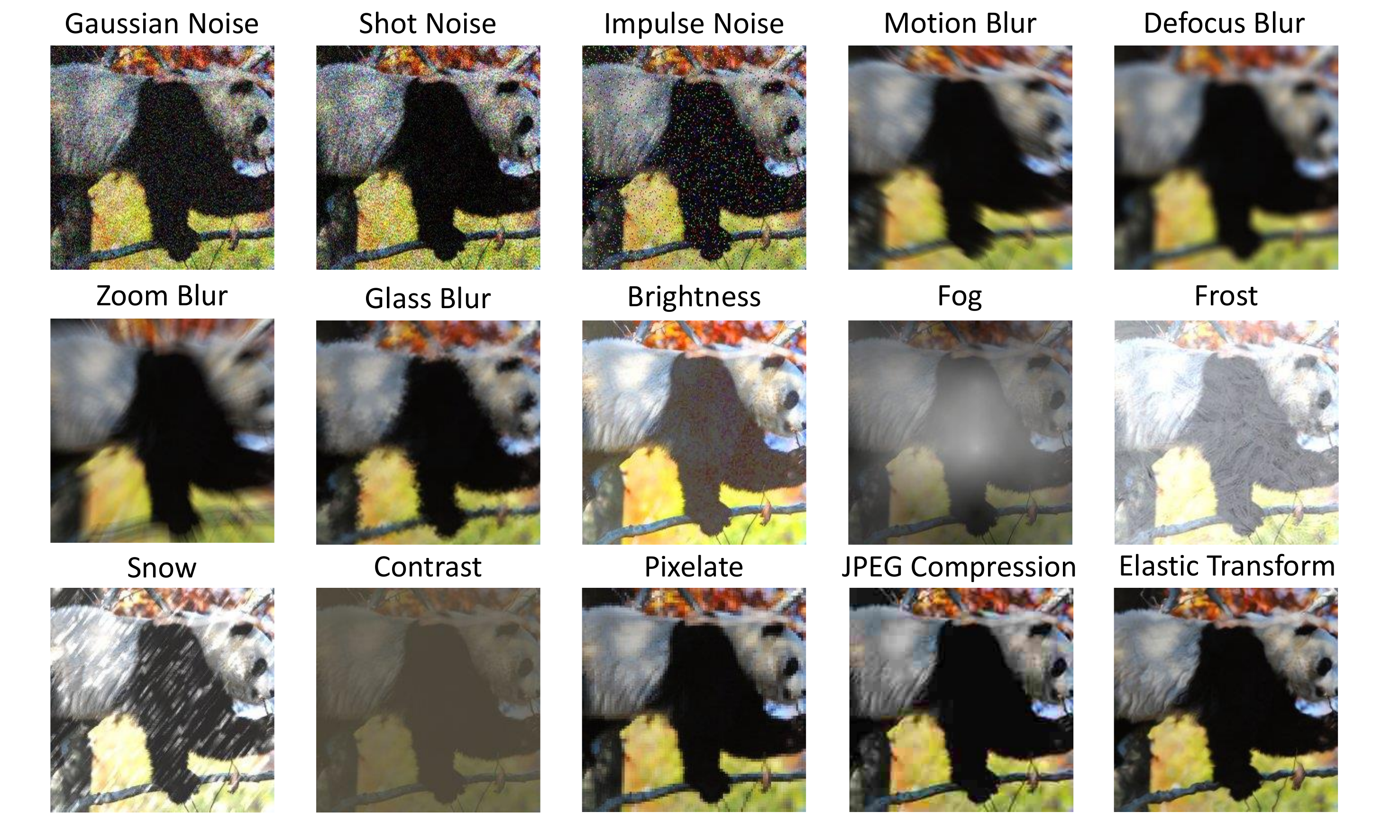}
   \caption{Examples of the 15 corruptions with severity level of 3 in the ImageNet-C corruption benchmark. }
   \label{fig:imagenetc}
\end{figure*}

The ImageNet-C benchmark consists of 15 types of algorithmically generated corruptions from noise (Gaussian noise, shot noise, impulse noise), blur (motion blur, defocus blur, zoom blur, glass blur), weather (brightness, fog, frost, snow), and digital categories (contrast, pixelate, JPEG compression, elastic transform) as shown in Fig. \ref{fig:imagenetc}.
The detailed description of these types of damage is as follows:

\begin{itemize}
    \item \textit{Gaussian noise} images are generated during the acquisition process.
          A digital device has inherent noise due to the lighting condition and its temperature.
    \item \textit{Shot noise}, also known as photon shot noise, is electronic noise caused by statistical quantum fluctuations, that is, variation in the number of photons sensed at a given exposure level.
    \item \textit{Impulse noise} is a color analog of salt-and-pepper noise and can be caused by analog-to-digital converter errors, bit errors in transmission, etc.
    \item \textit{Defocus noise} occurs when an image is out of focus.
    \item \textit{Glass blur} appears in photos taken through a frosted glass window.
    \item \textit{Motion blur} is the apparent streaking of moving objects in a photograph or a sequence of frames.
    \item \textit{Zoom blur} appears when a camera moves towards an object rapidly.
    \item \textit{Snow} type images are added with the effect of falling snow through an algorithm.
    \item \textit{Frost} type images are taken through frosted windows or lens.
    \item \textit{Fog} type images are generated with the diamond-square algorithm.
    \item \textit{Brightness} varies with light intensity.
    \item \textit{Contrast} can be high or low depending on the luminance or photographed object's color.
    \item \textit{Elastic transformations} stretch small regions.
    \item \textit{Pixelation} occurs when upsampling a low-resolution images.
    \item \textit{JPEG compression} is a commonly used method of lossy compression for digital images.
\end{itemize}

\begin{table}
    \centering
    \caption{AlexNet errors in ImageNet-C}
    \begin{tabular}{cc}
        \hline
        Corruption Type        & AlexNet Corruption Error \\
        \hline
        Gaussian Noise         & 88.6\%                   \\
        Shot Noise             & 89.4\%                   \\
        Impulse Noise          & 92.3\%                   \\
        Defocus Blur           & 82.0\%                   \\
        Glass Blur             & 82.6\%                   \\
        Motion Blur            & 78.6\%                   \\
        Zoom Blur              & 79.8\%                   \\
        Snow                   & 86.7\%                   \\
        Frost                  & 82.7\%                   \\
        Fog                    & 81.9\%                   \\
        Brightness             & 56.5\%                   \\
        Contrast               & 85.3\%                   \\
        Elastic Transformation & 64.6\%                   \\
        Pixelate               & 71.8\%                   \\
        JPEG Compression       & 60.7\%                   \\
        \hline
    \end{tabular}
    \label{tab:alexnet}
\end{table}

Since we use AlexNet errors to normalize Corruption Error values in ImageNet-C, we now specify the value $\frac{1}{5}\sum_{s=1}^5 E^\text{AlexNet}_{s,\text{Corruption}}$ for each corruption type in Tab. \ref{tab:alexnet}.

\section{Transferring Difference Directly}
In this section, we elaborate on how to obtain vicinal samples and how to directly use vicinal differences to improve robustness.
More importantly, we also conduct an ablation experiment using corrupted images for training.
This experiment strongly shows that transferring differences can make the networks insensitive to local changes in specific directions.

\begin{figure*}[h]
    \vspace{0cm}
    \centering
    \subfigure[Lack of vicinity sampling]{
        \label{fig:appendix_benefit_vita.1}
        \begin{minipage}[b]{0.30\textwidth}
            \includegraphics[width=1\textwidth]{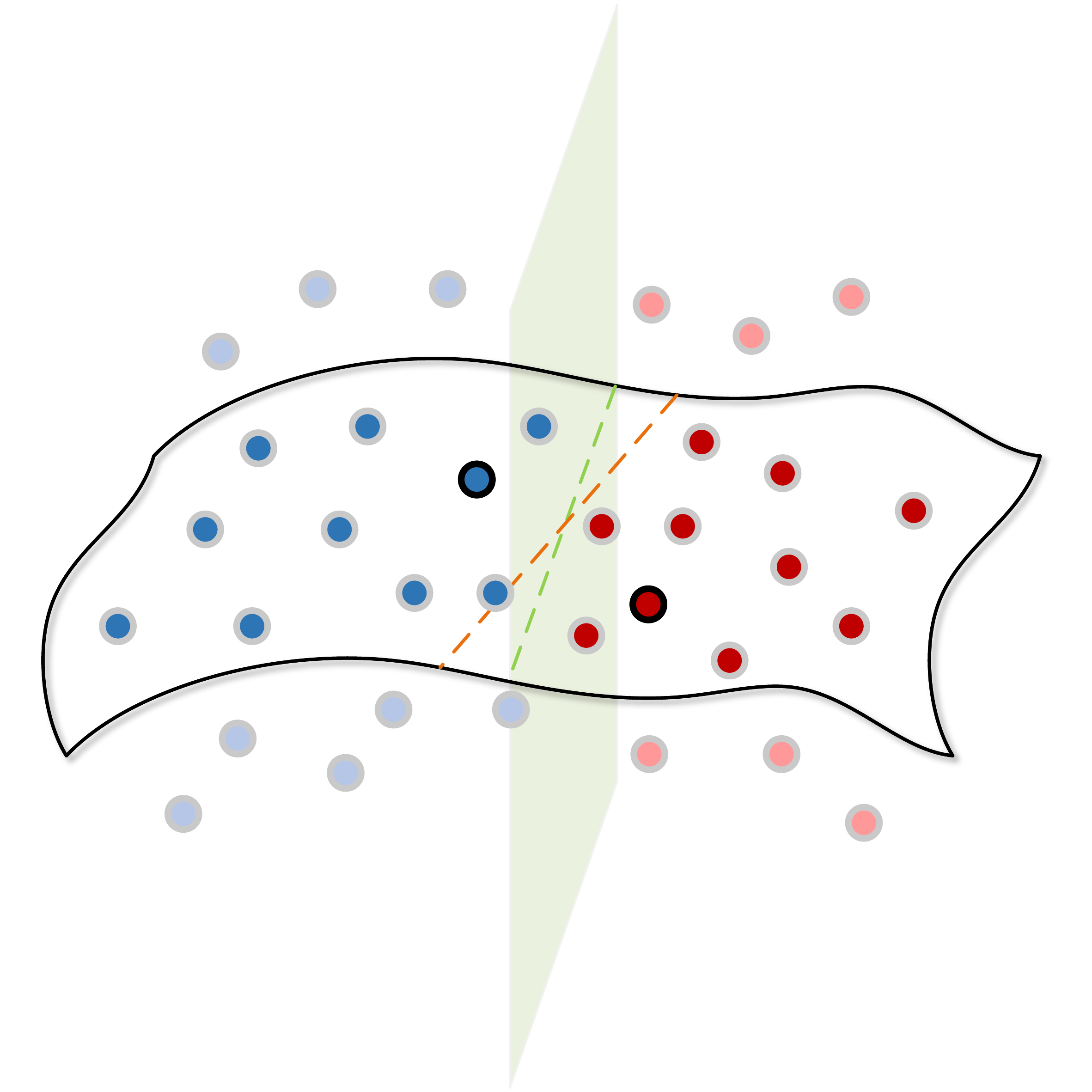}
        \end{minipage}}
    \hspace{8pt}
    \subfigure[Off-manifold samples hurts]{
        \label{fig:appendix_benefit_vita.2}
        \begin{minipage}[b]{0.30\textwidth}
            \includegraphics[width=1\textwidth]{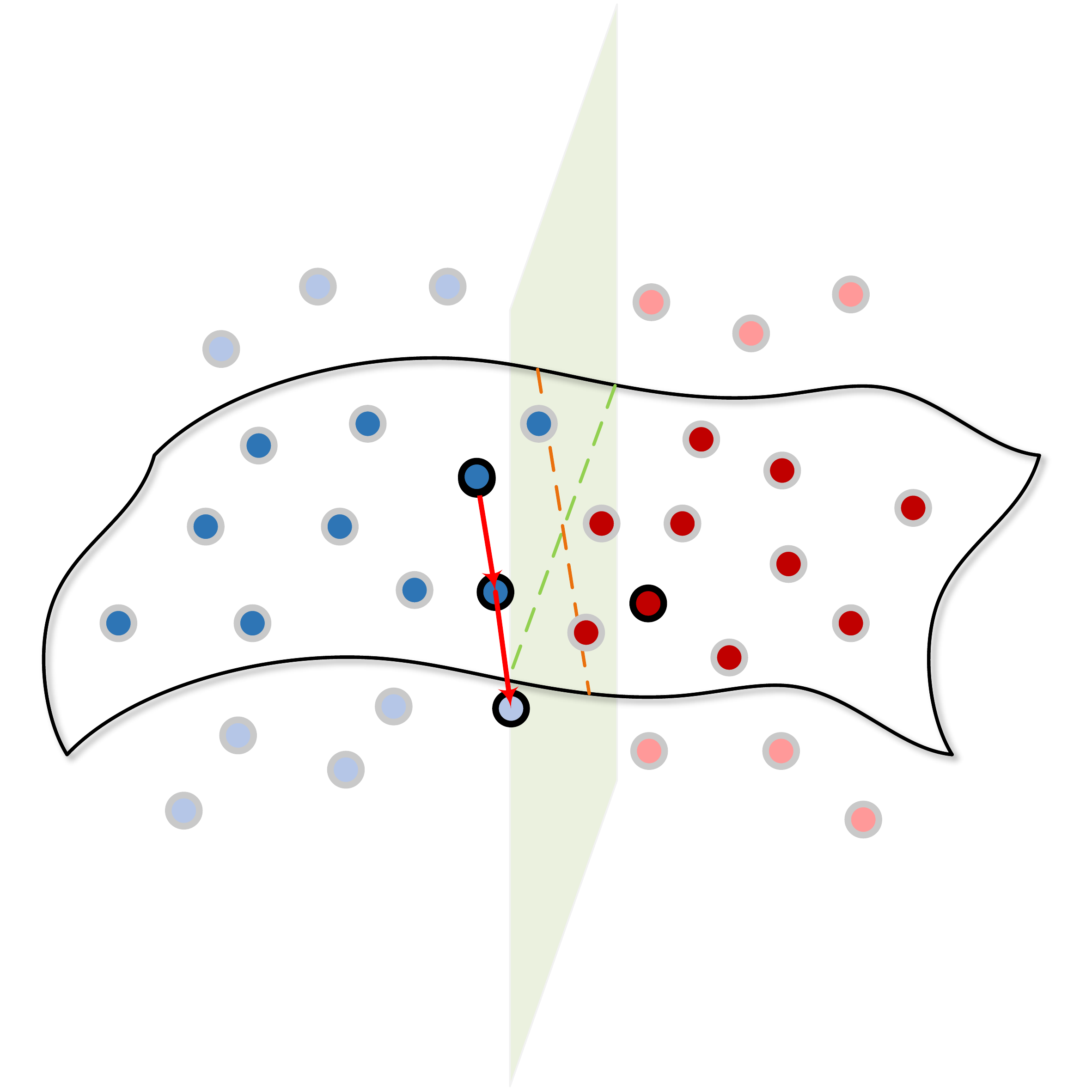}
        \end{minipage}}
    \hspace{8pt}
    \subfigure[Vicinal difference transfer]{
        \label{fig:appendix_benefit_vita.3}
        \begin{minipage}[b]{0.30\textwidth}
            \includegraphics[width=1\textwidth]{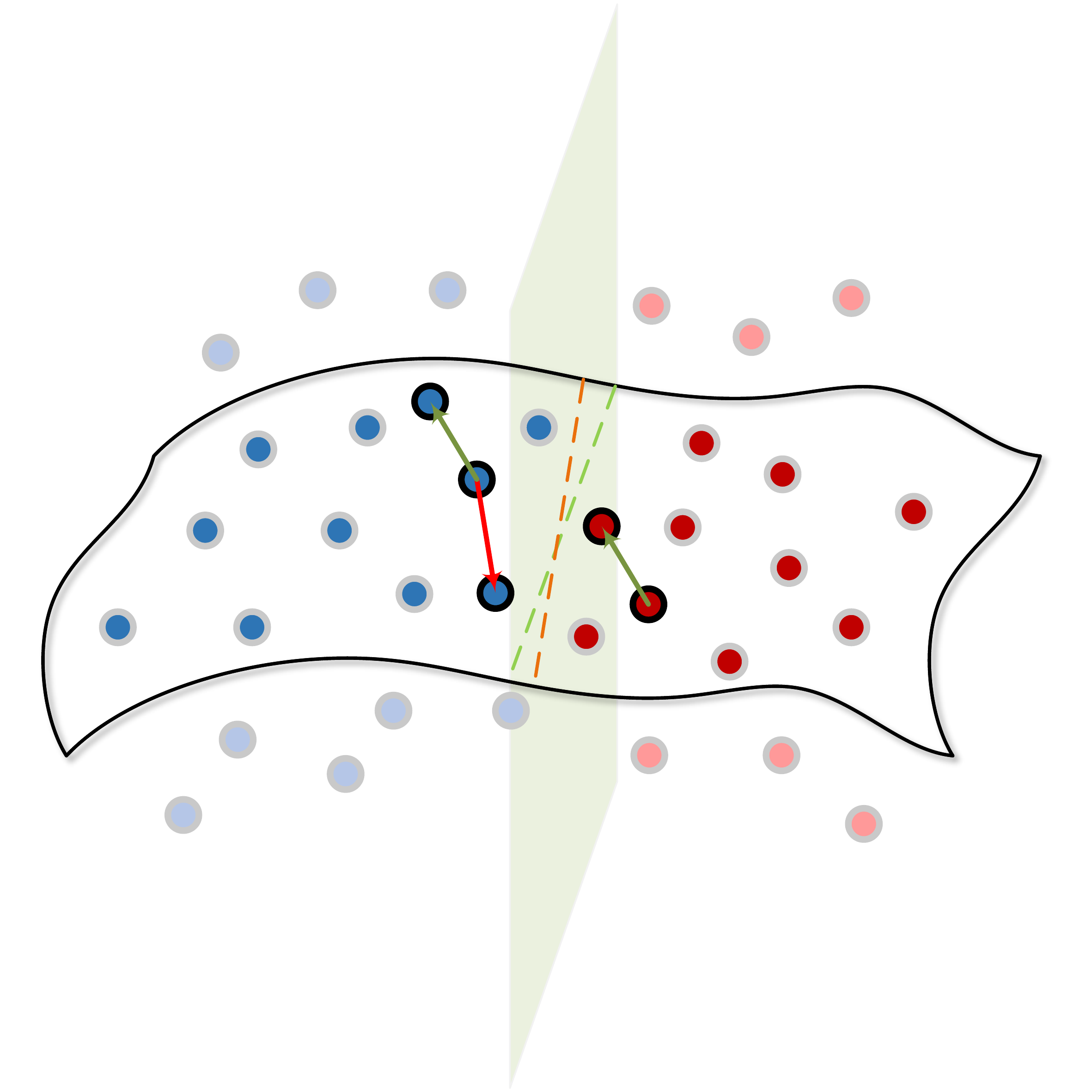}
        \end{minipage}}
    \caption{Three-dimensional schematic diagram to show that diverse on-manifold samples boost performance. \textbf{(a)} shows an underfitting phenomenon caused by the lack of sufficiently diverse vicinal samples. \textbf{(b)} takes a strong adversarial attack as an example to show the performance degradation caused by off-manifold samples. Classifiers tend to overfit these off-manifold samples and fail to construct a proper underlying manifold. \textbf{(c)} demonstrates the benefit of tangent transfer. While maintaining sample variety, it will not generate severe out-of-manifold samples. }
    \label{fig:appendix_benefit_vita}
    \vspace{0cm}
\end{figure*}

\subsection{Augmentation Source Details}
Since corruption datasets are used to measure model performance under data shifts, we carefully exclude the 15 corruptions into the VITA process and multi-source robust training.
The detailed description of each augmentation is as follows:
\begin{itemize}
    \item \textit{Rotation}. Rotate randomly from -5 to 5 degrees, and add a certain scaling operation to ensure that an augmented image can make a difference (\textit{i.e.} \textit{RandomAffine(degrees=5, scale=(0.95, 0.96))} in torchvision.transforms).
    \item \textit{Shearing}. Shear parallel to the $x$ or $y$ axis in range of (-2, 2) (\textit{i.e.} \textit{RandomAffine(shear=2, scale=(0.95, 0.96))} in torchvision.transforms).
    \item \textit{Translating}. Perform horizontal and vertical translations with maximum absolute fraction of 0.05 (\textit{i.e.} \textit{RandomAffine(translate=(0.05, 0.05), scale=(0.95, 0.96))} in torchvision.transforms).
    \item \textit{Cropping}. Crop 80\% to 85\% of the given image and resize it to a desired size (\textit{i.e.} \textit{RandomResizedCrop(size=ImgSize, scale=(0.8, 0.85))} in torchvision.transforms).
    \item \textit{Scaling}. Scale is randomly sampled from the range $1.05 \leq \text{scale} \leq 1.10$ (\textit{i.e.} \textit{RandomAffine(scale=(1.05, 1.10))} in torchvision.transforms).
    \item \textit{Solarize}. Solarize an image by inverting all pixel values above a threshold (\textit{i.e.} \textit{functional.solarize(img, threshold=128) in torchvision.transforms}).
    \item \textit{Posterize}. Posterize an image by reducing the number of bits for each color channel (\textit{i.e.} \textit{functional.posterize(img, bits=4) in torchvision.transforms}).
\end{itemize}

\subsection{Perturbation Source Details}
The adversarial attack methods we use all come from advertorch \cite{DBLP:journals/corr/abs-1902-07623}, which is a Python toolbox for adversarial robustness research.
The detailed description of each perturbation is as follows:
\begin{itemize}
    \item \textit{LinfPGDAttack}. PGD Attack \cite{DBLP:conf/iclr/MadryMSTV18} with order=Linf. The parameters are set as \textit{LinfPGDAttack(eps=0.03, nb\_iter=40, eps\_iter=0.001, rand\_init=True, targeted=False)}.
    \item \textit{L2PGDAttack}. PGD Attack with order=L2. The parameters are set as \textit{L2PGDAttack(eps=0.5, nb\_iter=40, eps\_iter=0.05, rand\_init=True, targeted=False)}.
    \item \textit{GradientSignAttack}. One step fast gradient sign method (Goodfellow \textit{et al, 2014} \cite{DBLP:journals/corr/GoodfellowSS14}). The parameters are set as \textit{GradientSignAttack(eps=0.05, targeted=False)}.
    \item \textit{LinfBasicIterativeAttack}. Like \textit{GradientSignAttack} but with several steps for each epsilon \cite{DBLP:conf/iclr/KurakinGB17}. The parameters are set as \textit{LinfBasicIterativeAttack(eps=0.03, nb\_iter=40, eps\_iter=0.001, targeted=False)}.
    \item \textit{L2BasicIterativeAttack}. Like \textit{GradientAttack} but with several steps for each epsilon. The parameters are set as \textit{L2BasicIterativeAttack(eps=1.0, nb\_iter=40, eps\_iter=0.05, targeted=False)}.
    \item \textit{MomentumIterativeAttack}. Adversarial attack with the momentum term into the iterative process for attacks \cite{DBLP:conf/cvpr/DongLPS0HL18}. The parameters are set as \textit{MomentumIterativeAttack(eps=0.03, nb\_iter=40, eps\_iter=0.001, targeted=False)}
    \item \textit{CarliniWagnerL2Attack}. The Carlini and Wagner L2 Attack \cite{DBLP:conf/sp/Carlini017}. The parameters are set as \textit{CarliniWagnerL2Attack(confidence=0.1, targeted=False, learning\_rate=0.01, binary\_search\_steps=9, max\_iterations=10, initial\_const=1e-3)}
\end{itemize}

\begin{figure*}[h]
    \vspace{-0.3cm}
    \centering
    \subfigure[Learned handbags data manifold]{
        \label{fig:vdt1}
        \begin{minipage}[b]{0.49\textwidth}
            \includegraphics[width=1\textwidth]{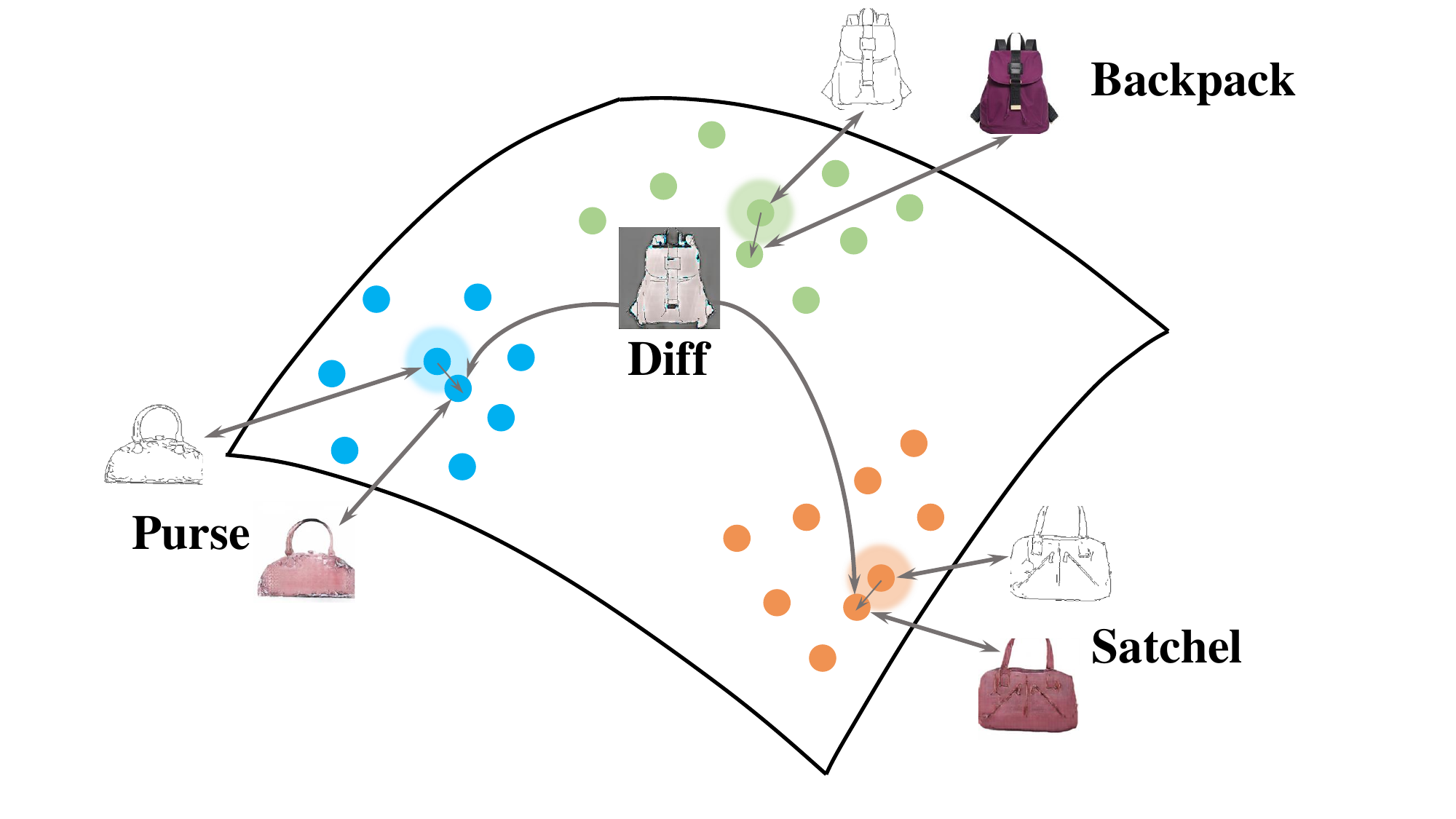}
        \end{minipage}}
    \subfigure[Learned CIFAR-10 data manifold]{
        \label{fig:vdt2}
        \begin{minipage}[b]{0.49\textwidth}
            \includegraphics[width=1\textwidth]{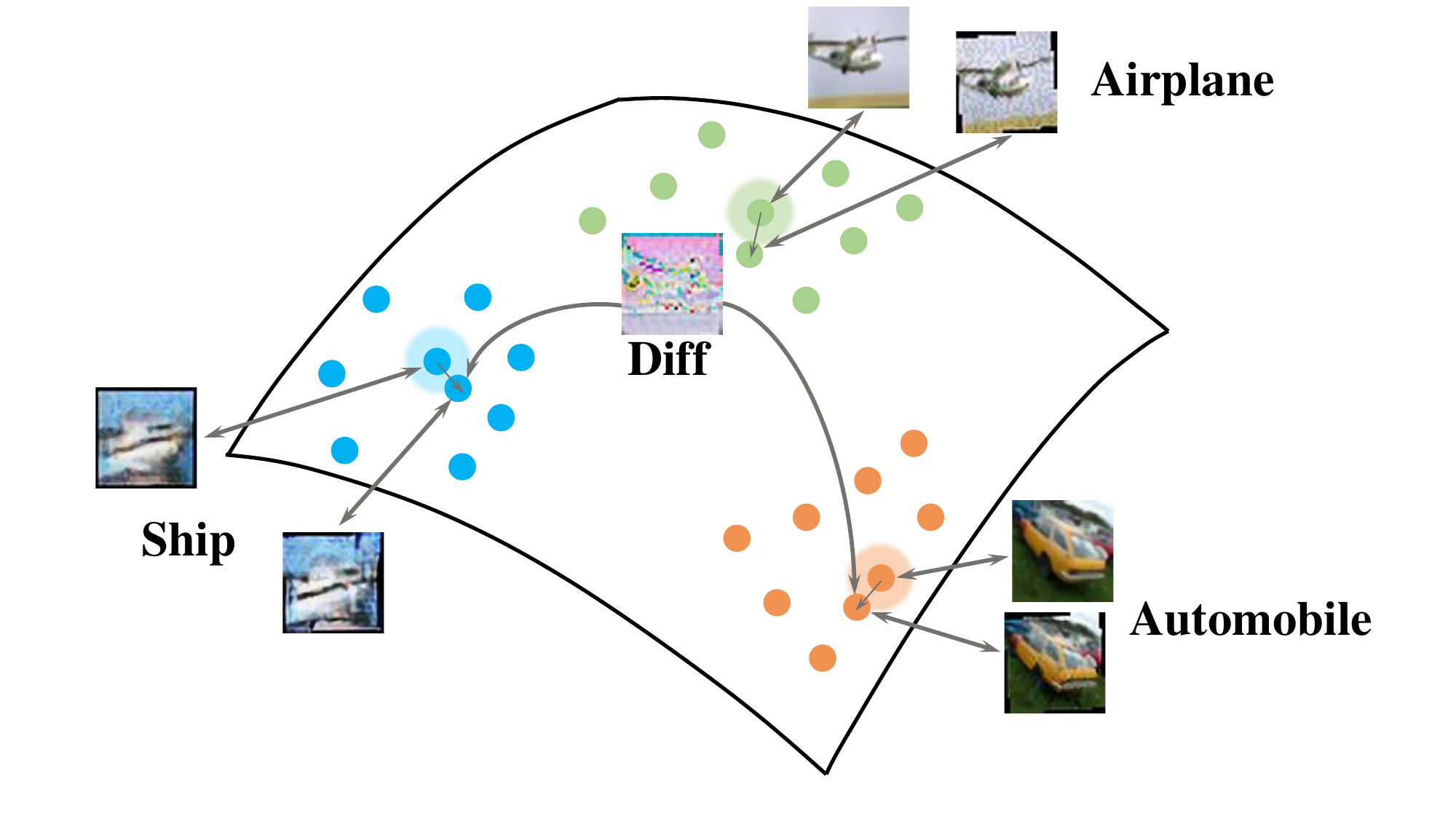}
        \end{minipage}}
    \caption{Shared difference on data manifold. \textbf{Left} images are from the edge2handbags dataset.
        Transferring the difference from the backpack, a colored purse or satchel can be crafted using our proposed generative model (for more detailed generated images see our supplementary).
        \textbf{Right} images are from the CIFAR-10 dataset.
        We utilize the vicinal difference between the original airplane and the augmented one to guide the generation of vicinal samples of other images.
        As shown on the right, a ship and a car can also be transformed as the airplane.}
    \label{fig:vdt}
    \vspace{-0.3cm}
\end{figure*}

\begin{figure*}
    \centering
    \includegraphics[width=\linewidth]{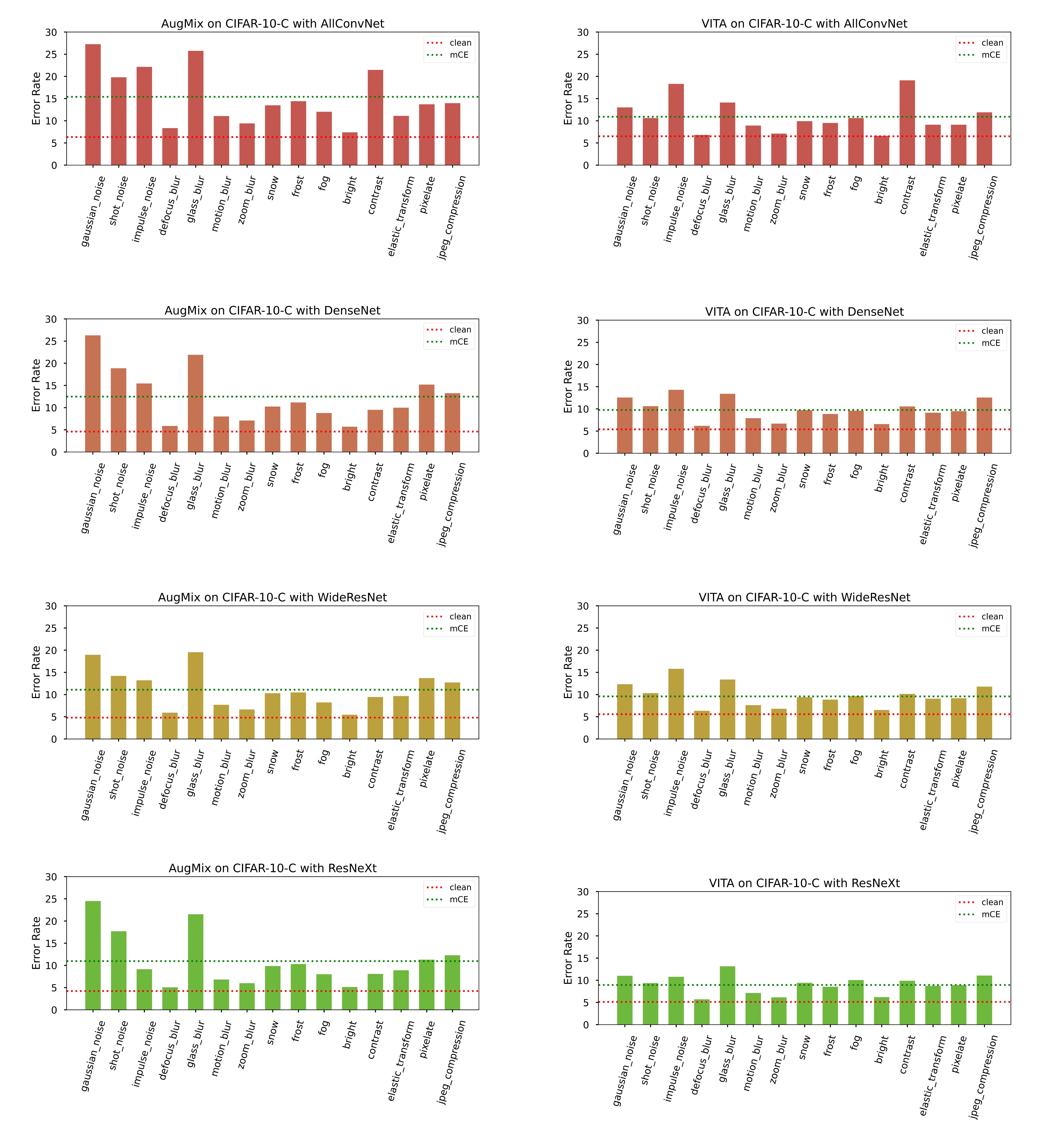}
    \caption{The excellent and balanced performance of VITA on Corruption datasets. It can be seen that even if a diverse augmentation strategy AugMix is adopted, the model will still have extremely uneven performance on different types of corrupted images. In contrast, VITA not only has significantly higher overall corruption robustness performance than AugMix, but also performs evenly towards different types of corruption. In particular, when VITA is used in an advanced network architecture such as ResNeXt, the performance gap between the best corruption type and the worst corruption type is less than 10\%, unlike AugMix, which has a gap of about 20\%.}
    \label{fig:augmix_bias}
\end{figure*}

\begin{figure*}
    \centering
    \includegraphics[scale=0.175]{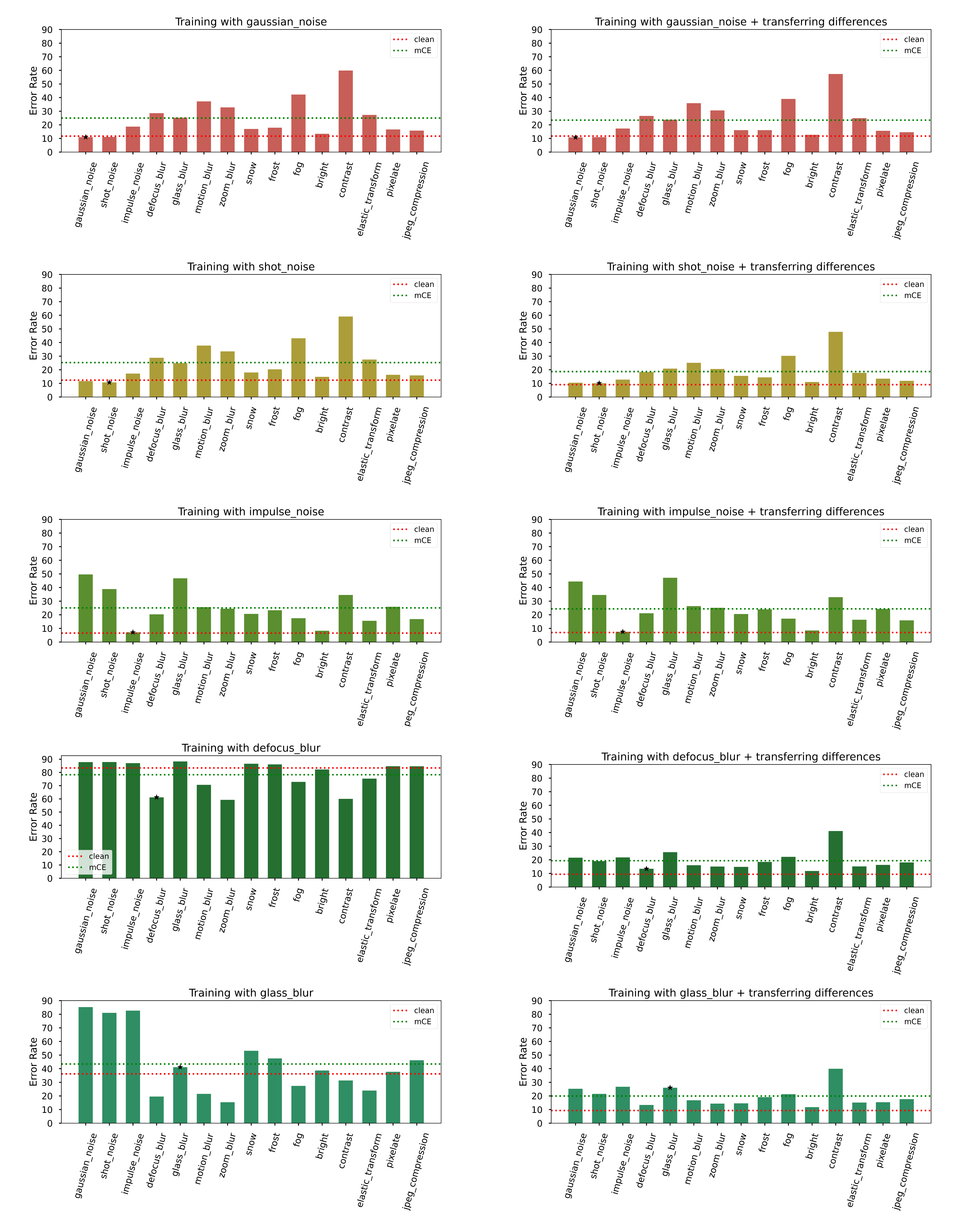}
    \caption{Training with corruption \textit{Gaussian noise}, \textit{shot noise}, \textit{impulse noise}, \textit{defocus blur} and \textit{glass blur}. }
    \label{fig:single_corruption_1}
\end{figure*}

\begin{figure*}
    \centering
    \includegraphics[scale=0.175]{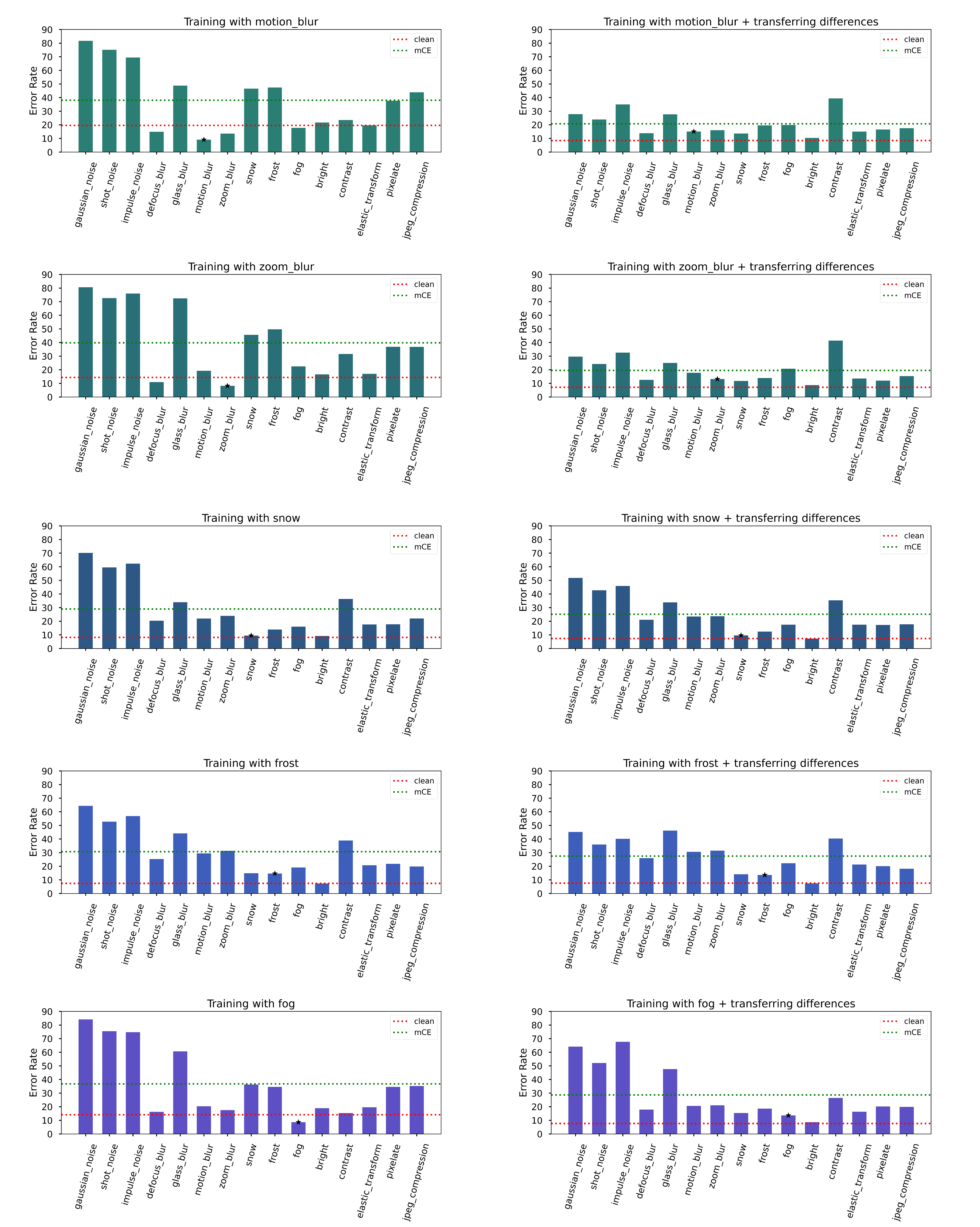}
    \caption{Training with corruption \textit{motion blur}, \textit{zoom blur}, \textit{snow}, \textit{frost} and \textit{fog}. }
    \label{fig:single_corruption_2}
\end{figure*}

\begin{figure*}
    \centering
    \includegraphics[scale=0.175]{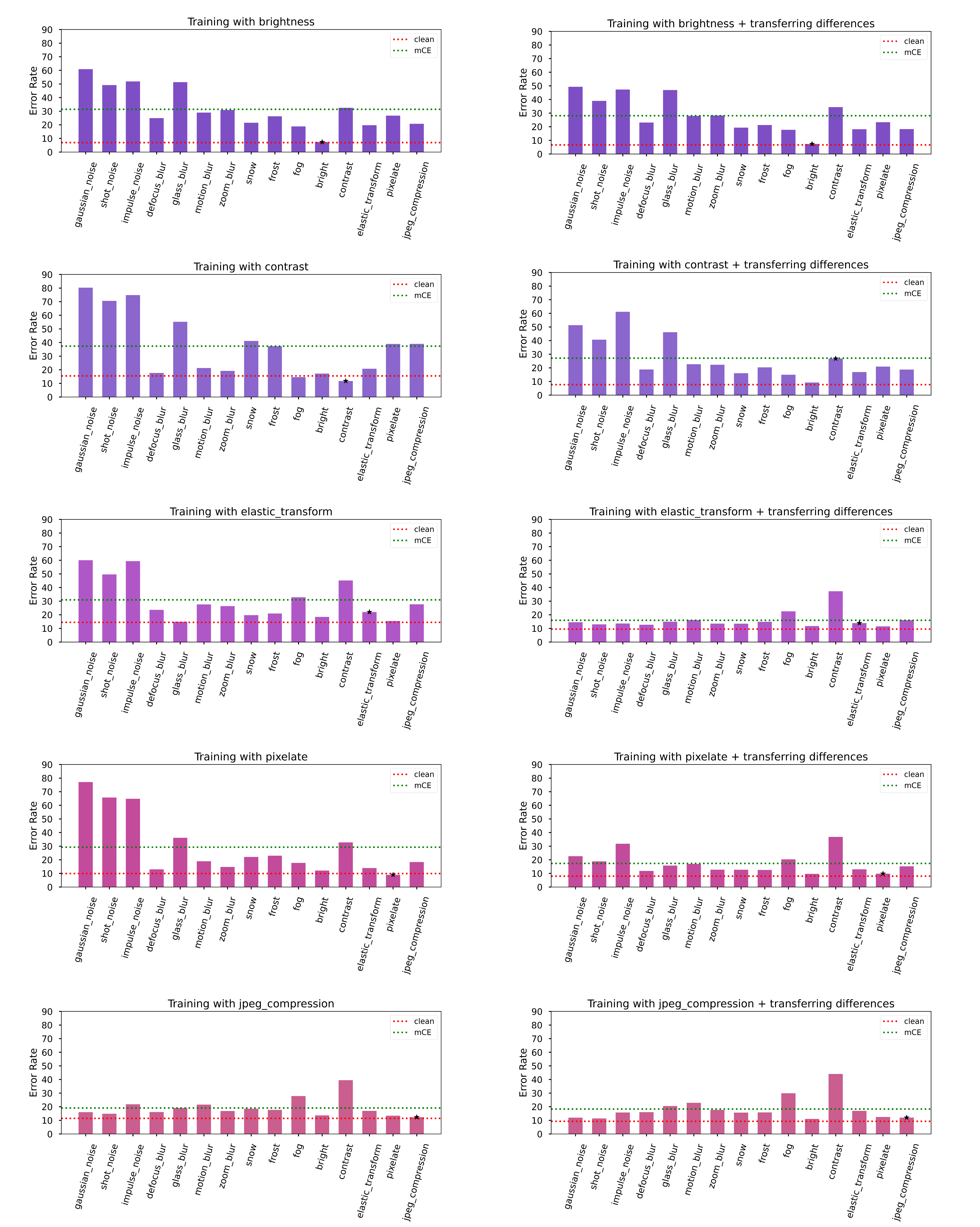}
    \caption{Training with corruption \textit{brightness}, \textit{contrast}, \textit{elastic transform}, \textit{pixelate} and \textit{JPEG compression}. }
    \label{fig:single_corruption_3}
\end{figure*}

\begin{figure*}
    \centering
    \includegraphics[scale=0.055]{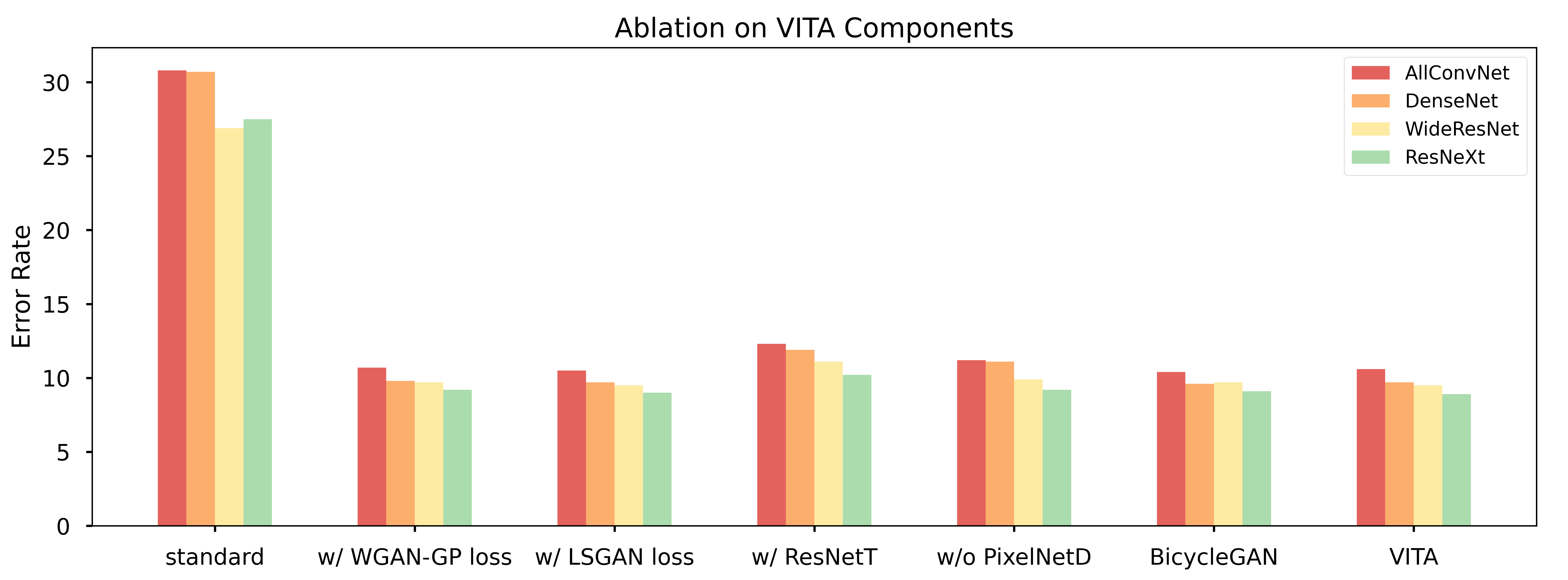}
    \caption{Ablation on VITA components. On the whole, the initial settings of VITA can get a good robust performance, and our generation framework is not very sensitive to the parameter settings of the components of the framework.}
    \label{fig:vita_ablation}
\end{figure*}

\renewcommand{\thesubfigure}{(\arabic{subfigure})}

\begin{figure*}[h]
    \centering
    \subfigure{
        \begin{minipage}[]{0.1\textwidth}
            \includegraphics[width=1\textwidth]{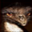} \\
            \includegraphics[width=1\textwidth]{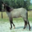} \\
            \includegraphics[width=1\textwidth]{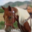} \\
            \includegraphics[width=1\textwidth]{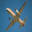} \\
            \includegraphics[width=1\textwidth]{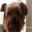} \\
            \includegraphics[width=1\textwidth]{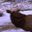}
            \vspace{-6mm}
            \caption*{(a)}
        \end{minipage}
        \begin{minipage}[]{0.1\textwidth}
            \includegraphics[width=1\textwidth]{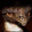} \\
            \includegraphics[width=1\textwidth]{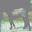} \\
            \includegraphics[width=1\textwidth]{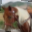} \\
            \includegraphics[width=1\textwidth]{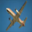} \\
            \includegraphics[width=1\textwidth]{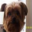} \\
            \includegraphics[width=1\textwidth]{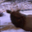}
            \vspace{-6mm}
            \caption*{(b)}
        \end{minipage}
        \begin{minipage}[]{0.1\textwidth}
            \includegraphics[width=1\textwidth]{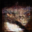} \\
            \includegraphics[width=1\textwidth]{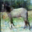} \\
            \includegraphics[width=1\textwidth]{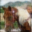} \\
            \includegraphics[width=1\textwidth]{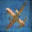} \\
            \includegraphics[width=1\textwidth]{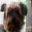} \\
            \includegraphics[width=1\textwidth]{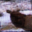}
            \vspace{-6mm}
            \caption*{(c)}
        \end{minipage}
    }
    \subfigure{
        \begin{minipage}[]{0.1\textwidth}
            \includegraphics[width=1\textwidth]{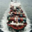} \\
            \includegraphics[width=1\textwidth]{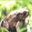} \\
            \includegraphics[width=1\textwidth]{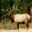} \\
            \includegraphics[width=1\textwidth]{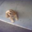} \\
            \includegraphics[width=1\textwidth]{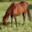} \\
            \includegraphics[width=1\textwidth]{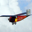}
            \vspace{-6mm}
            \caption*{(a)}
        \end{minipage}
        \begin{minipage}[]{0.1\textwidth}
            \includegraphics[width=1\textwidth]{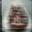} \\
            \includegraphics[width=1\textwidth]{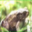} \\
            \includegraphics[width=1\textwidth]{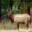} \\
            \includegraphics[width=1\textwidth]{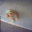} \\
            \includegraphics[width=1\textwidth]{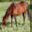} \\
            \includegraphics[width=1\textwidth]{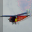}
            \vspace{-6mm}
            \caption*{(b)}
        \end{minipage}
        \begin{minipage}[]{0.1\textwidth}
            \includegraphics[width=1\textwidth]{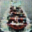} \\
            \includegraphics[width=1\textwidth]{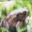} \\
            \includegraphics[width=1\textwidth]{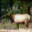} \\
            \includegraphics[width=1\textwidth]{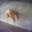} \\
            \includegraphics[width=1\textwidth]{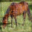} \\
            \includegraphics[width=1\textwidth]{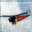}
            \vspace{-6mm}
            \caption*{(c)}
        \end{minipage}
    }
    \subfigure{
        \begin{minipage}[]{0.1\textwidth}
            \includegraphics[width=1\textwidth]{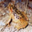} \\
            \includegraphics[width=1\textwidth]{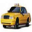} \\
            \includegraphics[width=1\textwidth]{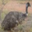} \\
            \includegraphics[width=1\textwidth]{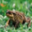} \\
            \includegraphics[width=1\textwidth]{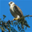} \\
            \includegraphics[width=1\textwidth]{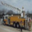}
            \vspace{-6mm}
            \caption*{(a)}
        \end{minipage}
        \begin{minipage}[]{0.1\textwidth}
            \includegraphics[width=1\textwidth]{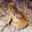} \\
            \includegraphics[width=1\textwidth]{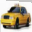} \\
            \includegraphics[width=1\textwidth]{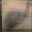} \\
            \includegraphics[width=1\textwidth]{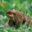} \\
            \includegraphics[width=1\textwidth]{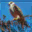} \\
            \includegraphics[width=1\textwidth]{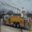}
            \vspace{-6mm}
            \caption*{(b)}
        \end{minipage}
        \begin{minipage}[]{0.1\textwidth}
            \includegraphics[width=1\textwidth]{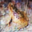} \\
            \includegraphics[width=1\textwidth]{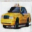} \\
            \includegraphics[width=1\textwidth]{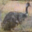} \\
            \includegraphics[width=1\textwidth]{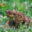} \\
            \includegraphics[width=1\textwidth]{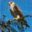} \\
            \includegraphics[width=1\textwidth]{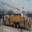}
            \vspace{-6mm}
            \caption*{(c)}
        \end{minipage}
    }
    \caption{Images from CIFAR-10 dataset. Here, (a) is the original image, (b) is the target image and (c) is the generated image.}
    \label{fig:cifar10}
    \vspace{0cm}
\end{figure*}

\renewcommand{\thesubfigure}{(\arabic{subfigure})}

\begin{figure*}[h]
    \centering
    \subfigure{
        \begin{minipage}[]{0.1\textwidth}
            \includegraphics[width=0.9\textwidth]{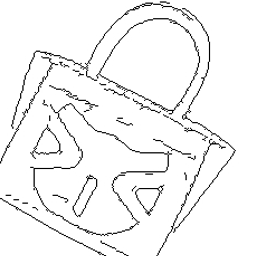} \\
            \includegraphics[width=0.9\textwidth]{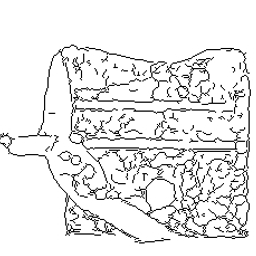} \\
            \includegraphics[width=0.9\textwidth]{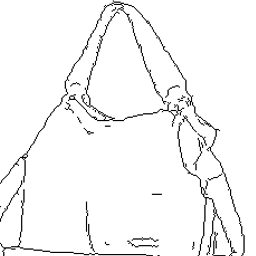} \\
            \includegraphics[width=0.9\textwidth]{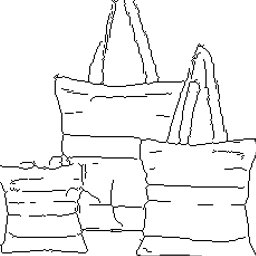} \\
            \includegraphics[width=0.9\textwidth]{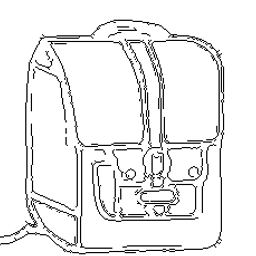}
            \vspace{0mm}
            \caption*{(a)}
        \end{minipage}
        \begin{minipage}[]{0.1\textwidth}
            \includegraphics[width=0.9\textwidth]{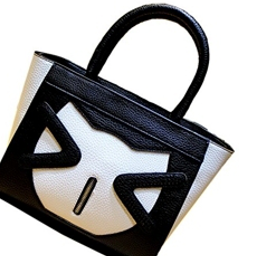} \\
            \includegraphics[width=0.9\textwidth]{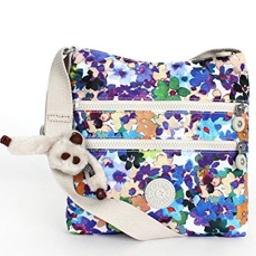} \\
            \includegraphics[width=0.9\textwidth]{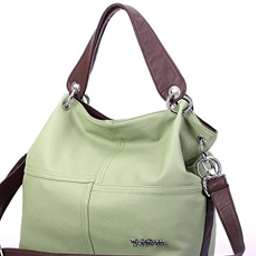} \\
            \includegraphics[width=0.9\textwidth]{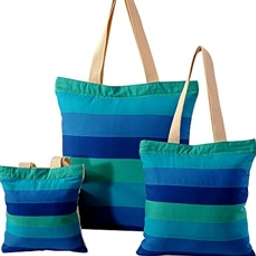} \\
            \includegraphics[width=0.9\textwidth]{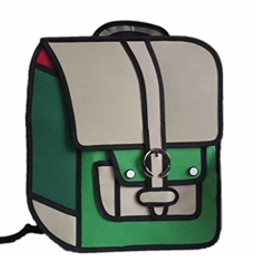}
            \vspace{0mm}
            \caption*{(b)}
        \end{minipage}
        \begin{minipage}[]{0.1\textwidth}
            \includegraphics[width=0.9\textwidth]{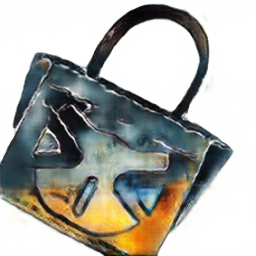} \\
            \includegraphics[width=0.9\textwidth]{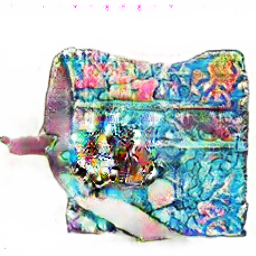} \\
            \includegraphics[width=0.9\textwidth]{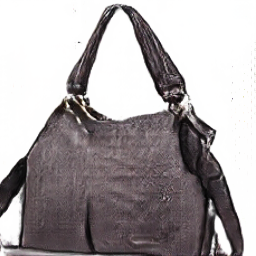} \\
            \includegraphics[width=0.9\textwidth]{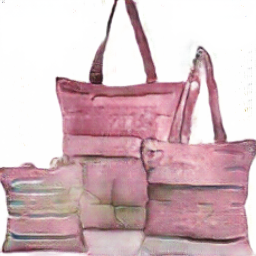} \\
            \includegraphics[width=0.9\textwidth]{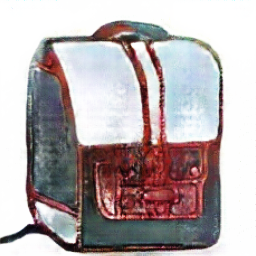}
            \vspace{0mm}
            \caption*{(c)}
        \end{minipage}
        \begin{minipage}[]{0.1\textwidth}
            \includegraphics[width=0.9\textwidth]{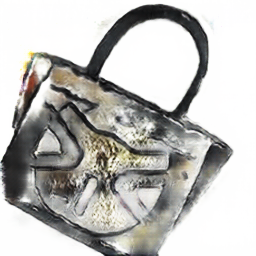} \\
            \includegraphics[width=0.9\textwidth]{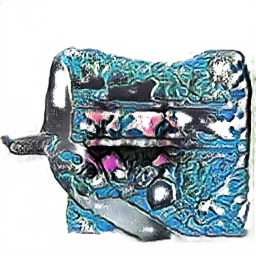} \\
            \includegraphics[width=0.9\textwidth]{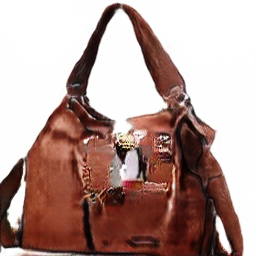} \\
            \includegraphics[width=0.9\textwidth]{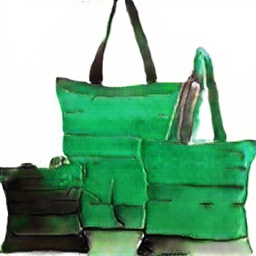} \\
            \includegraphics[width=0.9\textwidth]{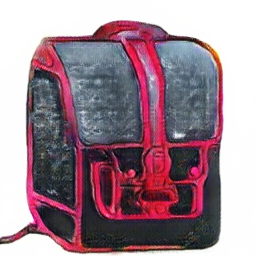}
            \vspace{0mm}
            \caption*{(d)}
        \end{minipage}
    }
    \subfigure{
        \begin{minipage}[]{0.1\textwidth}
            \includegraphics[width=0.9\textwidth]{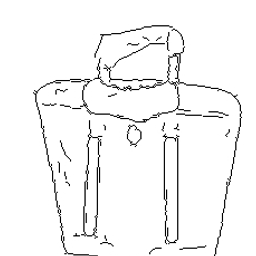} \\
            \includegraphics[width=0.9\textwidth]{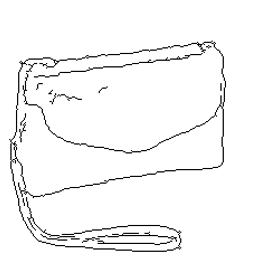} \\
            \includegraphics[width=0.9\textwidth]{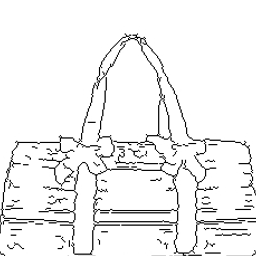} \\
            \includegraphics[width=0.9\textwidth]{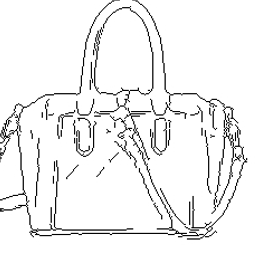} \\
            \includegraphics[width=0.9\textwidth]{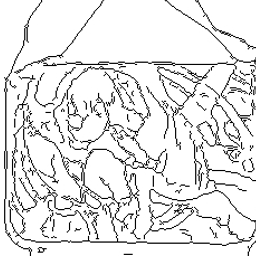}
            \vspace{0mm}
            \caption*{(a)}
        \end{minipage}
        \begin{minipage}[]{0.1\textwidth}
            \includegraphics[width=0.9\textwidth]{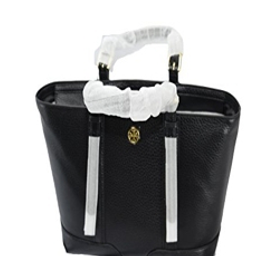} \\
            \includegraphics[width=0.9\textwidth]{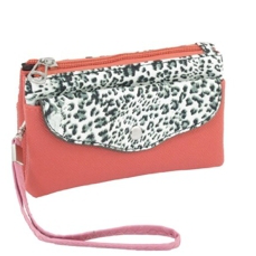} \\
            \includegraphics[width=0.9\textwidth]{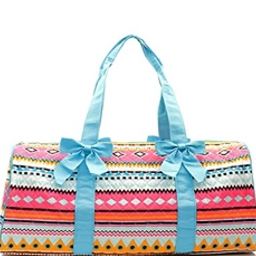} \\
            \includegraphics[width=0.9\textwidth]{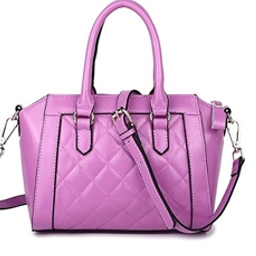} \\
            \includegraphics[width=0.9\textwidth]{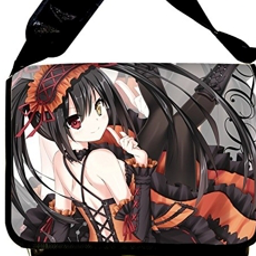}
            \vspace{0mm}
            \caption*{(b)}
        \end{minipage}
        \begin{minipage}[]{0.1\textwidth}
            \includegraphics[width=0.9\textwidth]{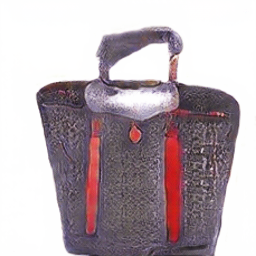} \\
            \includegraphics[width=0.9\textwidth]{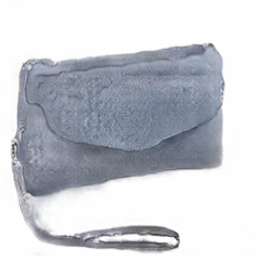} \\
            \includegraphics[width=0.9\textwidth]{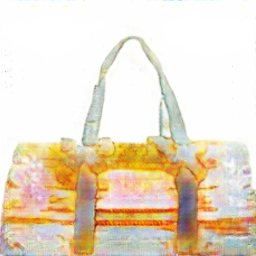} \\
            \includegraphics[width=0.9\textwidth]{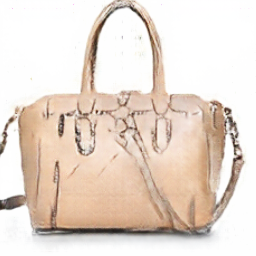} \\
            \includegraphics[width=0.9\textwidth]{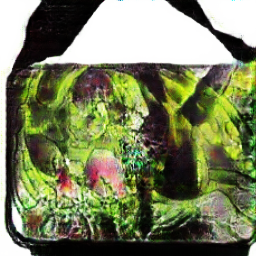}
            \vspace{0mm}
            \caption*{(c)}
        \end{minipage}
        \begin{minipage}[]{0.1\textwidth}
            \includegraphics[width=0.9\textwidth]{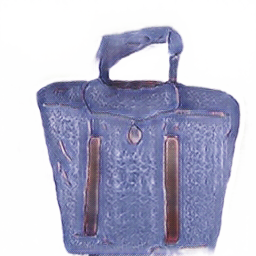} \\
            \includegraphics[width=0.9\textwidth]{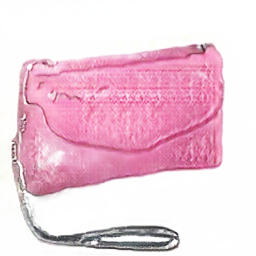} \\
            \includegraphics[width=0.9\textwidth]{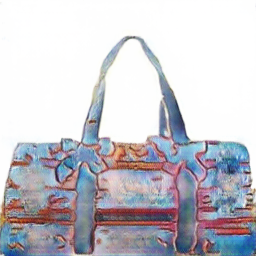} \\
            \includegraphics[width=0.9\textwidth]{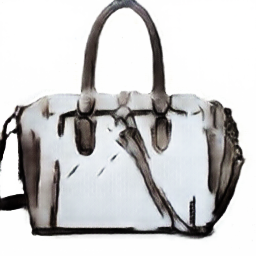} \\
            \includegraphics[width=0.9\textwidth]{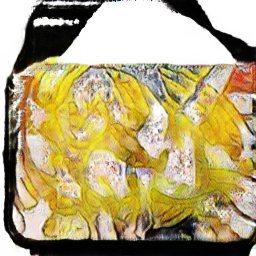}
            \vspace{0mm}
            \caption*{(d)}
        \end{minipage}
    }
    \caption{Images from \textit{edges2handbags} dataset. Here, (a) is the input (original) image, (b) is the target (ground truth) image, (c) is one generated image from VITA and (d) is another generated image from VITA}
    \label{fig:edges2handbags}
\end{figure*}

\begin{table*}[tp]
   \centering
   \small
   \caption{
      Evaluations of adversarial robustness (test accuracy) of Wide ResNet on the CIFAR-10 dataset.
      It can be seen that our method not only improves the robustness but also improves the accuracy of clean samples significantly.
      Here, $\epsilon_\text{train}$ is the maximum value of adversarial perturbation during training ($\epsilon_\text{test} = 8 / 255$ same standard adversarial robustness evaluation). In standard adversarial training, performance on clean samples hurt with larger $\epsilon_\text{train}$. In contrast, a model trained with VITA via multi-source integrated training can have a better performance on clean samples.}
   \vspace{5pt}
   \label{tab:more_adv}
   \renewcommand\arraystretch{1.0}
   \begin{tabular}{c|c|p{1.5cm}<{\centering}|p{1.5cm}<{\centering}|p{1.5cm}<{\centering}|p{2.0cm}<{\centering}}
      \hline
      \textbf{Network Architectures} & \textbf{Method}                                           & \textbf{Clean} & \textbf{FGSM}  & \textbf{PGD-20} & \textbf{C\&W ($l_{\infty}$)} \\
      \hline
      \hline
      \multirow{4}{*}{WRN-58-10}     & FAT for TRADES ($\epsilon_\text{train} = 8/255$)          & 87.09          & 68.70          & 57.17           & 55.43                        \\
                                     & VITA for Adv. Training ($\epsilon_\text{train} = 8/255$)  & \textbf{88.42} & \textbf{73.05} & 60.55           & 55.87                        \\
                                     & FAT for TRADES ($\epsilon_\text{train} = 16/255$)         & 85.28          & 68.08          & 58.39           & 55.89                        \\
                                     & VITA for Adv. Training ($\epsilon_\text{train} = 16/255$) & 86.85          & 72.76          & \textbf{60.83}  & \textbf{56.77}               \\
      \hline
   \end{tabular}
\end{table*}

\begin{table*}[tp]
    \centering
    \caption{Contributions of different data sources to improve the corruption robustness (AllConvNet on CIFAR-10-C). Without samples generated by our method, the corruption robustness of the model is poor, which reflects the irreplaceability of our generated samples to significant improvement on the corruption robustness. Besides, only using any one or two sources of data for training does not have the effect of using all sources of data. This shows the necessity of multi-source integrated training.}
    \label{tab:source_ablation_detail}
    \renewcommand\arraystretch{1.0}
    \begin{tabular}{c||c||c|c|c|c|c|c}
        \hline
        \textbf{Corruption Type} & \textbf{All Source} & \textbf{Only Aug} & \textbf{Only Adv} & \textbf{Only Gen} & \textbf{w/o Aug} & \textbf{w/o Adv} & \textbf{w/o Gen} \\
        \hline
        \hline
        % Clean &                    & 6.4            & 9.5          & 9.9          & 6.5                    & \textbf{5.7}         & 6.9                  \\

        Clean                    & 6.5                 & 6.4               & 9.5               & 9.9               & 6.5              & \textbf{5.6}     & 6.9              \\
        \hline
        \hline
        Noise                    & 14.0                & 20.9              & 41.9              & 21.1              & \textbf{14.8}    & 18.2             & 18.1             \\
        \hline
        Gaussian Noise           & 13.0                & 24.4              & 42.9              & 23.0              & \textbf{15.2}    & 21.3             & 18.6             \\
        Shot Noise               & 10.6                & 18.1              & 35.7              & 18.5              & \textbf{12.3}    & 15.5             & 14.1             \\
        Impulse Noise            & 18.3                & 20.1              & 47.0              & 21.9              & \textbf{16.9}    & 17.9             & 21.5             \\

        \hline
        \hline
        Blur                     & 9.2                 & 12.2              & 18.3              & 15.1              & \textbf{10.2}    & 10.3             & 11.8             \\
        \hline
        Defocus Blur             & 6.8                 & 7.8               & 12.9              & 11.6              & 7.8              & \textbf{6.8}     & 8.7              \\
        Glass Blur               & 14.1                & 21.9              & 27.2              & 21.8              & \textbf{14.5}    & 17.7             & 17.0             \\
        Motion Blur              & 8.9                 & 10.3              & 19.7              & 14.0              & 10.3             & \textbf{9.0}     & 12.0             \\
        Zoom Blur                & 7.1                 & 8.9               & 13.5              & 12.8              & 8.3              & \textbf{7.3}     & 9.4              \\
        \hline
        \hline

        Weather                  & 9.2                 & 11.1              & 16.9              & 11.8              & 10.1             & \textbf{9.2}     & 11.8             \\
        \hline
        Snow                     & 9.9                 & 12.9              & 17.2              & 12.7              & 11.3             & \textbf{10.6}    & 12.7             \\
        Frost                    & 9.5                 & 13.1              & 16.4              & 12.1              & \textbf{9.7}     & 10.4             & 12.1             \\
        Fog                      & 10.6                & 10.9              & 23.0              & 13.9              & 12.1             & \textbf{9.5}     & 13.9             \\
        Brightness               & 6.6                 & 7.5               & 11.1              & 8.3               & 7.3              & \textbf{6.4}     & 8.3              \\
        \hline
        \hline

        Digital                  & 12.3                & 14.7              & 21.3              & 15.2              & 12.5             & \textbf{11.7}    & 15.2             \\
        \hline
        Contrast                 & 19.1                & 21.0              & 39.3              & 26.8              & 20.4             & \textbf{16.3}    & 26.8             \\
        Elastic                  & 9.1                 & 10.6              & 15.2              & 11.1              & 10.1             & \textbf{9.1}     & 11.1             \\
        Pixelate                 & 9.1                 & 13.5              & 16.3              & 11.2              & \textbf{8.9}     & 10.1             & 11.2             \\
        JPEG                     & 11.9                & 13.5              & 14.2              & 11.8              & \textbf{10.7}    & 11.2             & 11.8             \\

        \hline
        \hline
        Mean                     & 10.9                & 14.3              & 23.4              & 16.6              & \textbf{11.7}    & 11.9             & 13.9             \\
        \hline
    \end{tabular}
\end{table*}

\subsection{Ablation Study on Augmentation Types}
Here, through ablation experiments, we further show that transferring differences is critical to alleviating the model's sensitivity to certain local changes along the sample manifold.
For this reason, we directly add the corruption type in the corruption datasets during the training to compare the improvement on corruption robustness of the model after transferring differences.
It should be noted that our above training process will adopt the Jensen-Shannon regularization term.
To further illustrate the effectiveness of transferring differences and reflect the tendency of the model trained with one type of corruption (severity is 1) to a specific local direction, we do not use the regularization term as a comparison in the training process here.

From Fig. \ref{fig:single_corruption_1}, \ref{fig:single_corruption_2} and \ref{fig:single_corruption_3}, a model trained with specific corruption type has an obvious tendency on corruption robustness, that is, perform extremely well on some corruption types (usually similar to the corruption types added in training), and perform very poor on some of the other types.
While the model undergoes transferring differences during the training process, it can be seen that not only the overall performance (both on clean and corrupted images) of the model has been improved, but also the performance of various types of corruption is more balanced.

Additionally, if the difference produced by an augmentation is basically irrelevant to the original image (such as Gaussian noise), transferring differences have little impact on performance.
With the same augmentation strategy, samples will have similar local variation, but this local variation will have deviations between different samples.
We argue that the deviation of transferred differences introduces more kinds of vicinal information.

\section{VITA: Vicinal Transfer Augmentation}

In this section, we first show our generated samples from \textit{edges2handbags} and the CIFAR-10 dataset.
Then we demonstrate that VITA performs better and balanced under various corruption types compared to AugMix \cite{DBLP:conf/iclr/HendrycksMCZGL20}.
Finally, we conduct ablation experiments on the main components of our generative framework.
Besides, since our target samples are weakly enhanced samples and adversarial samples, we can see that the samples we generated in CIFAR are not much different from the original images.
But our generated samples have a great impact on the corruption robustness of the model.
Therefore, we argue that current data augmentation research needs to focus more on the design of refined weak augmentation strategies.

\subsection{Translator Generated Images}

Here, we show generated images, including images from \textit{edges2handbags} (Fig. \ref{fig:edges2handbags}) and CIFAR-10 dataset (Fig. \ref{fig:cifar10}).
These generated images show that our generative framework is also suitable for image translation tasks, and can also generate diverse outputs.

\subsection{Balanced Performance of VITA}

From Fig. \ref{fig:single_corruption_1}, \ref{fig:single_corruption_2} and \ref{fig:single_corruption_3}, we can find that the performance of a model trained with a certain type of augmentation will be extremely unbalanced under different corruption.
Further, we show in Fig. \ref{fig:augmix_bias} that even if a model trained with diverse augmentation (e.g. AugMix), will also have obvious imbalance performance on different types of corruption.
As a comparison, our augmentation method VITA can not only significantly outperform AugMix in overall performance but also has a highly balanced performance under various types of corruption.

\subsection{Ablation Study on VITA Components}
We have done ablation experiments on our generation method from four levels, namely loss function, translator, discriminator, and overall framework.

Our initial loss for GAN is the cross-entropy objective used in the original GAN \cite{DBLP:conf/nips/GoodfellowPMXWOCB14}.
From Fig. \ref{fig:vita_ablation}, we find that using the loss from WGAP-GP \cite{DBLP:conf/nips/GulrajaniAADC17} or LSGAN \cite{DBLP:conf/iccv/MaoLXLWS17} can bring a certain performance improvement, but the improvement is not large.

In terms of the translator, we use the UNet \cite{DBLP:conf/miccai/RonnebergerFB15} architecture by default, to better retain the difference information.
Here, we replace UNet with the converter of the ResNet \cite{DBLP:conf/eccv/HeZRS16} architecture (\textit{i.e.} w/ ResNetT), and find that the effect is indeed not as good as UNet, but the performance degradation is not serious.

In the default setting in our experiment, our discriminator is PatchGAN discriminator \cite{DBLP:conf/cvpr/IsolaZZE17}.
For comparison, we use a normal convolutional network as the discriminator instead and find that the performance of the model trained with samples from our method is slightly reduced.

Our generation method is mainly based on the pix2pix framework \cite{DBLP:conf/cvpr/IsolaZZE17}.
Here, we desire to know whether using a more complex image translation framework can bring further performance improvements.
From the Fig. \ref{fig:vita_ablation}, we can see that the model trained with the samples generated by BicycleGAN \cite{DBLP:conf/nips/ZhuZPDEWS17} has a certain increase in corruption robustness, but the improvement is not obvious.

In general, our generation method is not sensitive to specific hyper-parameters.

\section{Multi-source Robust Training}
In this part, we supplement related details on multi-source robust training.
First, we describe the specific implementation in detail.
Then, we display the performance of multi-source robust training with different sources towards different types of corruption.
Finally, we show the performance of multi-source robust training for adversarial robustness in a larger model.

\subsection{Training Details}

The key to our multi-source integration training is to have multi-source samples in a batch, so that the statistical parameters of the batch normalization layer will not be biased towards the distribution of specific sources.
Since the samples generated by VITA contain the transferred difference information, it is an optional option whether weakly augmented samples and adversarial examples need to undergo transferring difference in our Algorithm \ref{alg:code}.
This has little impact on the final performance, and our default setting in the experiment is to perform difference transfer.

\subsection{More on Data Source Ablation}
In Tab. \ref{tab:source_ablation_detail}, we show in detail how different data sources contribute to corruption robustness.
It further strengthens the irreplaceable role of the samples generated from VITA to improve the corruption robustness of models.
Meanwhile, it can be seen that whether it is a single source of data or any combination of two sources, their performance is not as good as the integration of data from all sources, which reflects the necessity of our multi-source robust training.

\subsection{More on Adversarial Training}

We employ Wide ResNet with a larger capacity (\textit{i.e.} WRN-58-10) to show the advanced performance of multi-source integrated on improving adversarial robustness.
It can be seen from the Tab. \ref{tab:more_adv} that our method has a more significant improvement effect on a larger network, especially the performance on PGD-20.

\end{document}